\pdfoutput=1

\documentclass[11pt]{article}

\usepackage[final]{acl}

\usepackage{times}
\usepackage{latexsym}
\usepackage{booktabs}
\usepackage{amsmath}

\usepackage[T1]{fontenc}

\usepackage[utf8]{inputenc}

\usepackage{microtype}

\usepackage{inconsolata}

\usepackage{graphicx}
\usepackage{float}
\usepackage{placeins}
\usepackage{enumitem}

\title{Understanding In-Context Learning Beyond Transformers: An Investigation of State Space and Hybrid Architectures}

\author{Shenran Wang \And Timothy Tin-Long Tse \And Jian Zhu \\\AND
        \parbox{\textwidth}{
            \normalfont
            \centering
            The University of British Columbia \\
            \texttt{shenranw@cs.ubc.ca}, \texttt{ttse05@student.ubc.ca}, \texttt{jian.zhu@ubc.ca}
        }
         }

\begin{document}
\maketitle
\begin{abstract}
We perform in-depth evaluations of in-context learning (ICL) on state-of-the-art transformer, state-space, and hybrid large language models over two categories of knowledge-based ICL tasks. Using a combination of behavioral probing and intervention-based methods, we have discovered that, while LLMs of different architectures can behave similarly in task performance, their internals could remain different. We discover that function vectors (FVs) responsible for ICL are primarily located in the self-attention and Mamba layers, and speculate that Mamba2 uses a different mechanism from FVs to perform ICL. FVs are more important for ICL involving parametric knowledge retrieval, but not for contextual knowledge understanding. Our work contributes to a more nuanced understanding across architectures and task types. Methodologically, our approach also highlights the importance of combining both behavioural and mechanistic analyses to investigate LLM capabilities.
\end{abstract}

\section{Introduction}
In-context learning (ICL) \citep{icl, llms-few-shot-learners}, the ability to learn from few-shot demonstrations at test time, is an emergent ability from pretrained Large Language Models (LLMs). In ICL, a few training samples are presented as demonstrations in the prompt, from which the model can learn to make predictions without any parameter updates. This emergent ability has raised interest in research on how LLMs acquire and generalize patterns solely from contexts at test time.

Mechanistic interpretability studies have successfully attributed ICL in transformers to certain types of self-attention heads, notably FVs (FVs) \cite{function-vectors,icl-fv,icl-heads} and induction heads \cite{olsson2022context,edelman2024evolution,icl-heads}, but such analyses are limited to transformers \cite{vaswani2017attention}.

\begin{figure}[H]
    \centering
    \includegraphics[width=\linewidth]{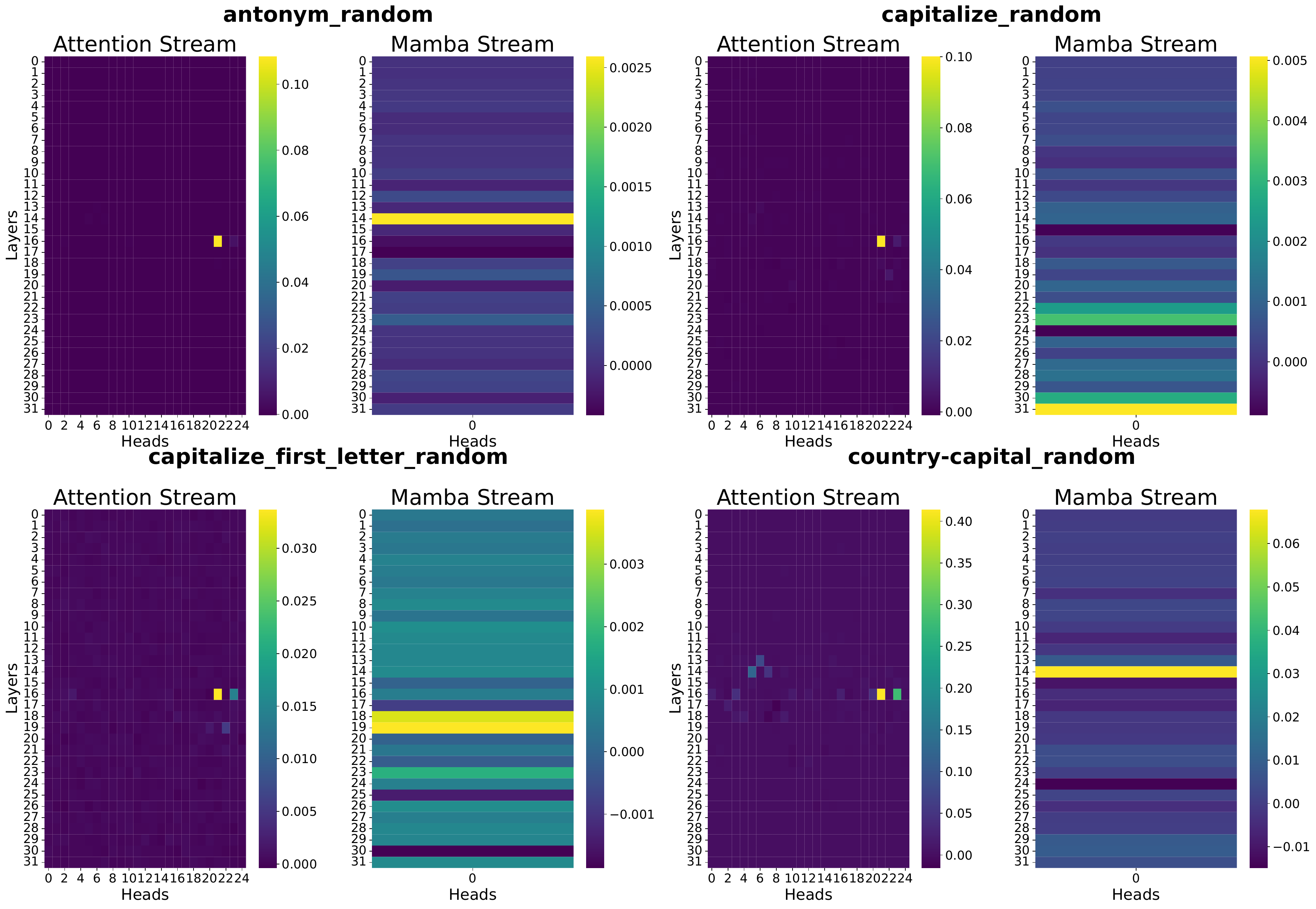}
    \caption{\textsc{Hymba-1.5B-Base}'s AIE heatmap on a subset of parametric knowledge retrieval ICL. Top FV heads identified are much more concentrated in self-attention layers than in SSM layers. [X-axis: head number; Y-axis: layer number.]}
    \label{fig:heatmap-fv_og-hymba}
\end{figure}
\begin{figure}[H]
    \centering
    \includegraphics[width=\linewidth]{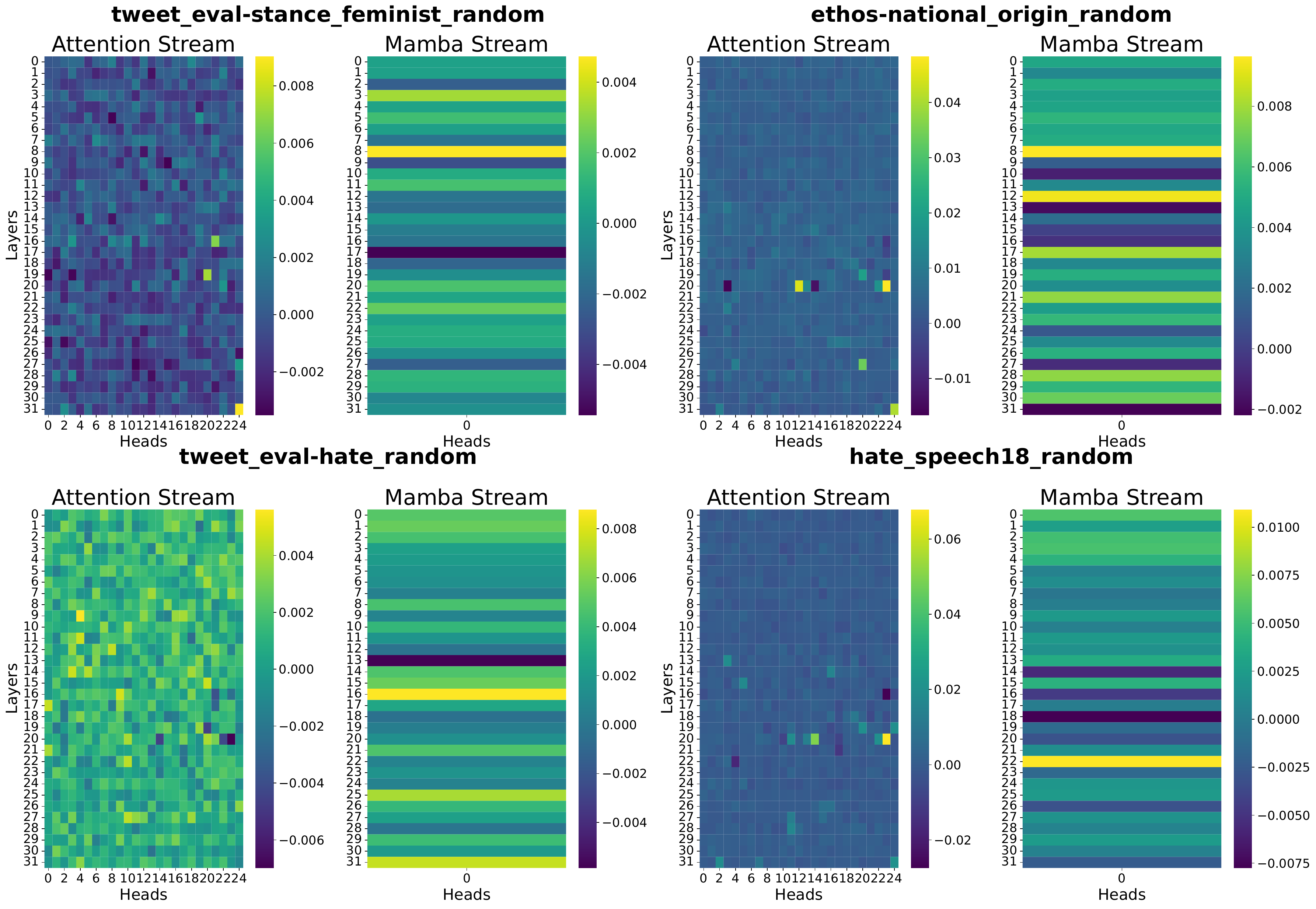}
    \caption{\textsc{Hymba-1.5B-Base}'s AIE heatmap on a subset of contextual knowledge understanding ICL. The top FV heads are far less concentrated than in parametric knowledge retrieval ICL. [X-axis: head number; Y-axis: layer number.]}
    \label{fig:heatmap-classification-hymba}
\end{figure}

Some early explorations \cite{grazzi2024mamba,park2024can,li2024can,li2025understanding,bondaschi2025markov} show that state-space models (SSMs) like Mamba \cite{mamba} and Mamba2 \cite{mamba2} can also perform ICL, though their ICL capabilities are weaker than transformers. 
Transformer-SSM hybrid architectures are therefore proposed to improve the ICL in SSMs \cite{park2024can}. However, little is known about the internal mechanisms of ICL in SSMs and hybrid models.

To address this gap, we analyze ICL in transformers, state space models, and hybrid models. Built on prior works \cite{rethink-demonstrations,function-vectors,icl-heads}, we further distinguish two types of knowledge-based ICL tasks that require different reasoning abilities \cite{massive-values}. A combination of behaviroal and mechanistic interpretability experiments yield several novel observations not previously reported in the literature.
\begin{itemize}[noitemsep, topsep=0pt, left=1em]
    \item While all architectures demonstrate qualitatively similar ICL performance and robustness against noisy labels, mechanistic interpretability analyses reveal differences in internal mechanisms. 
    \item FVs contribute more to the ICL capabilities in transformers, Mamba, and hybrid models, but less to Mamba2. 
    \item For hybrid models, ICL capabilities are primarily contributed by the FVs located in self-attention layers, regardless of whether the self-attention layers and SSM layers are stacked in parallel or interleaved sequantially. 
    \item FVs play a more important role in parametric knowledge retrieval ICL tasks. For ICL tasks that require contextual understanding, FVs have less impact on performance. These two types of ICL tasks do not always share the same set of FVs.
\end{itemize}
Our work has extended the prior analysis of FVs \cite{function-vectors,icl-fv,icl-heads} to more diverse architectures and further contributes to a more nuanced understanding of FVs in different knowledge-based task types. Methodologically, our approach also highlights the importance of combining both behavioural and mechanistic analyses to investigate model behaviors. Our code is available at: \texttt{\url{https://github.com/ShenranTomWang/ICL}}.

\section{Related Work}
\paragraph{Mamba and Hybrid LLMs} Mamba \cite{mamba} and Mamba2 \cite{mamba2} are state-space models (SSMs) with selective state update mechanisms. Mamba-based and Mamba-transformer hybrid LLMs have also emerged as more efficient replacements for transformer-based LLMs. Zamba \cite{zamba}, Zamba2 \cite{zamba2}, Jamba \cite{jamba}, Samba \cite{samba}, and Nemotron-H \cite{nemotron-h} are representative hybrid models that stack self-attention and Mamba/Mamba2 sequentially. Hymba \cite{hymba} is a hybrid model that integrates self-attention and Mamba in parallel. In this research, we examine the ICL capabilities of these state space and hybrid models, which have remained understudied.

\paragraph{In-Context Learning} Since first observed by \citet{llms-few-shot-learners}, ICL has been studied across different perspectives. \citet{what-can-icl} and \citet{transformers-linear-icl} provide theoretical insights into tasks that transformer models can learn in-context. \citet{icl-implicit-bayesian}, \citet{icl-learnability}, \citet{llms-implicit-topic-models} explores the theory behind ICL. \citet{rethink-demonstrations}, \cite{more-than-prompt}, and \citet{ground-truth-matter} studies how prompts can affect ICL performance. Other than transformer models, \citet{can-mamba-icl} studies the performance of Mamba-based models in ICL. \citet{pretrain-icl} and \citet{meta-icl} propose to improve ICL performance of LLMs in training time. \citet{function-vectors} and \citet{which-fv-head} examine the ICL mechanism through the lens of mechanistic interpretability. In our research, we will focus on hybrid models, and perform analysis on both the prompt level and the model internals.

\paragraph{Function Vectors and Function Vector Heads} Recent work by \citet{icl-fv} and \citet{function-vectors} has demonstrated that certain attention heads in LLMs are responsible for in-context learning, namely function vector (FV) heads. \citet{which-fv-head} further studies the roles and importance of FV induction heads in in-context learning, identifying how effective each function head is in LLMs. However, FVs have only been identified in transformer architectures, and little is known about the internal mechanism of ICL in non-transformer architectures. To fill in this gap, we will apply similar mechanistic interpretability techniques to investigate the FVs in state-space models and hybrid architectures.

\section{Models and Tasks}
To understand ICL, we selected diverse but representative architectures and knowledge-based tasks. 

\begin{figure*}[t]
    \centering
    \includegraphics[width=0.48\linewidth]{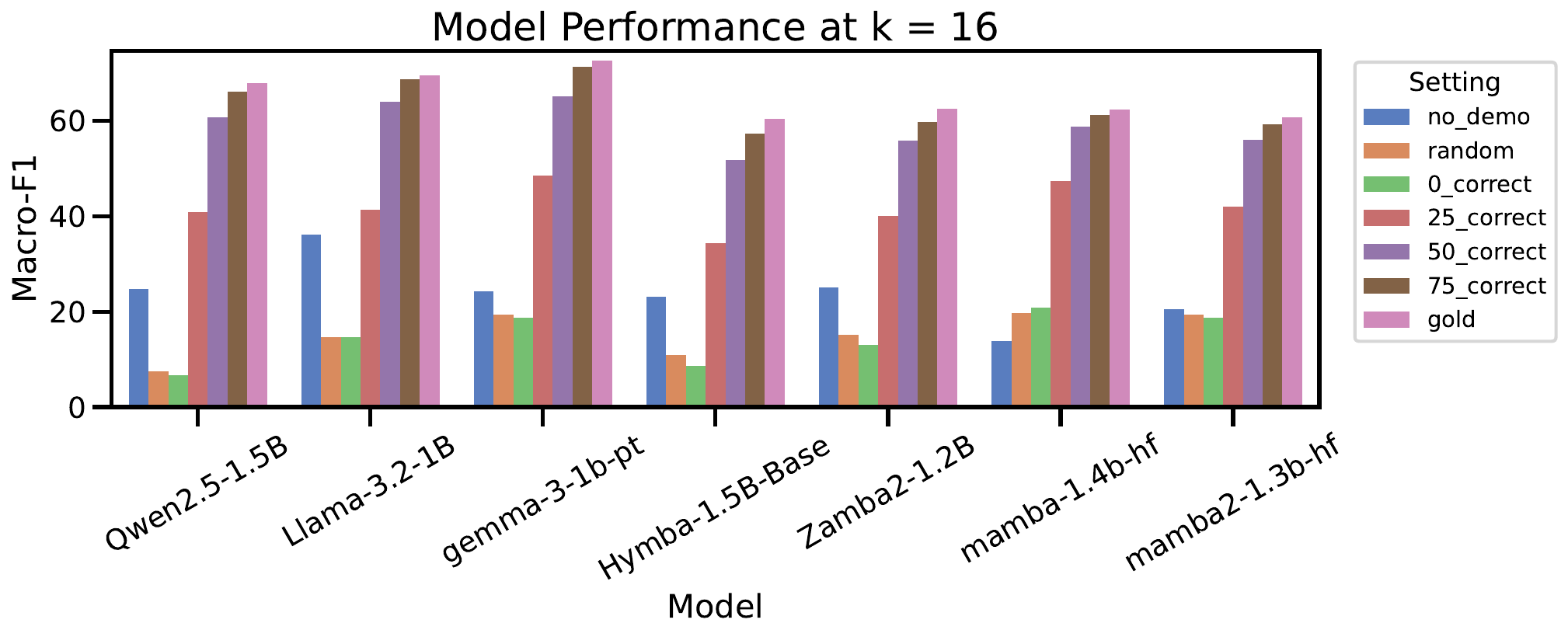}
    \includegraphics[width=0.48\linewidth]{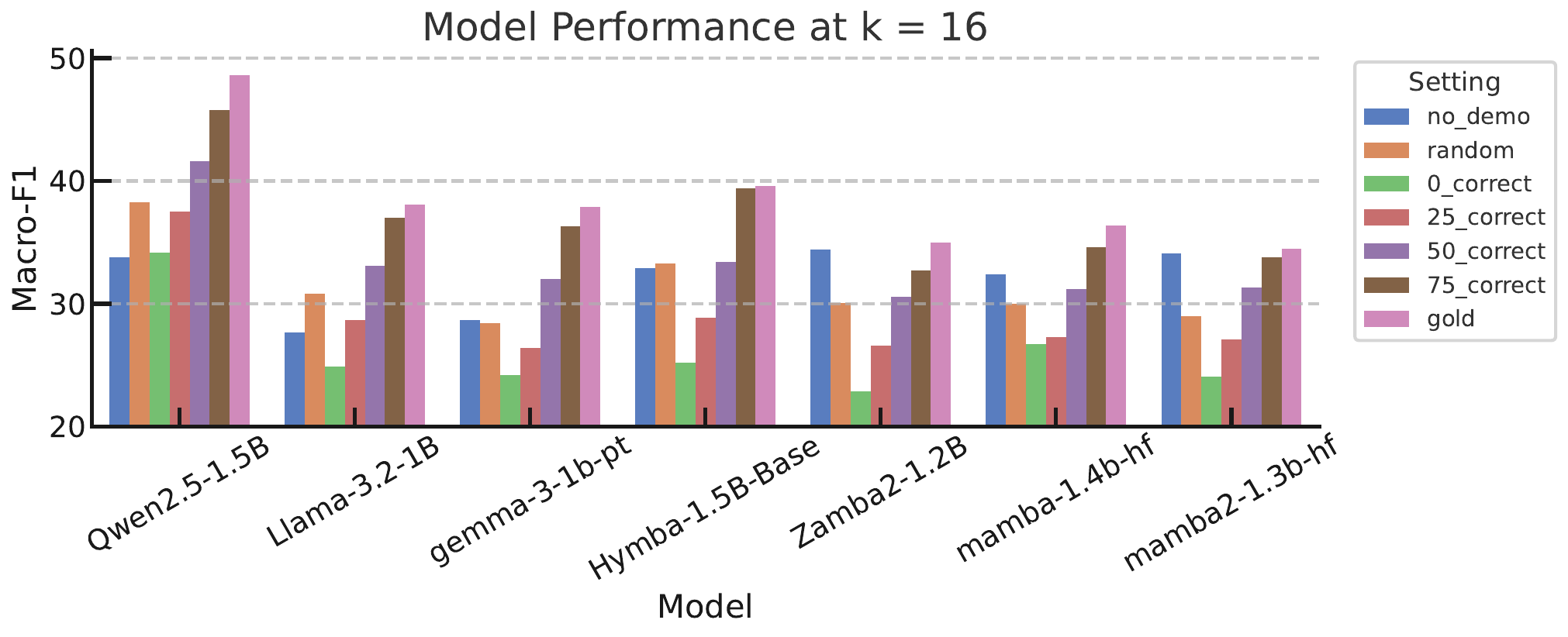}
    \caption{\textbf{Left:} Performance of different models on parametric knowledge retrieval datasets, for $k$ = 16, on initial setting. \textbf{Right:} Performance of different models on contextual knowledge understanding datasets, for $k$ = 16, on initial setting. }
    \label{fig:k=16}
\end{figure*}
\subsection{Baseline models}
\label{sec:baseline_models}
For SSMs, we chose pretrained checkpoints of \textsc{Mamba-1.4B}\footnote{\url{https://huggingface.co/state-spaces/mamba-1.4b-hf}} \cite{mamba}, \textsc{Mamba2-1.3B}\footnote{\url{https://huggingface.co/AntonV/mamba2-1.3b-hf}} \cite{mamba2}. For pretrained transformer LLMs, \textsc{gemma-3-1b-pt} \cite{gemma-3}, \textsc{Llama-3.2-1B} \cite{llama-3.2} and \textsc{Qwen2.5-1.5B} \cite{qwen-2.5} were selected as baselines. All models were limited to around 1B parameters for fair comparison.

\subsection{Hybrid Models}
\label{sec:hybrid_models}
We selected \textsc{Hymba-1.5B-Base} \cite{hymba} and \textsc{Zamba2-1.2B} \cite{zamba2}. They were chosen becasue they had competitive performance with transformer LLMs and they represent two typical designs of hybrid models, that is, parallel and sequentially-stacked hybrid layers.

\paragraph{Zamba2} In \textsc{Zamba2}, a self-attention layer with LoRA \cite{lora} is stacked on top of every 6 Mamba2 layers. A linear layer is then applied to the outputs of the self-attention layer, followed by more Mamba2 layers. Our selected \textsc{Zamba2-1.2B} has 37 Mamba2 layers, each with 64 heads, and 6 self-attention layers, each with 32 attention heads.

\paragraph{Hymba} Unlike in \textsc{Zamba2}, \textsc{Hymba} incorporates Mamba and self-attention blocks in parallel. At each layer, inputs individually go through Mamba and sliding-window self-attention blocks and are taken the mean before going through the out projection. Our selected \textsc{Hymba-1.5B-Base} has 32 layers. Each layer has 25 self-attention heads and 1 Mamba head.

\subsection{Tasks}
\label{sec:tasks}
Adopting the definitions in \citet{massive-values}, we classified our ICL tasks into \textbf{Contextual Knowledge Understanding} and \textbf{Parametric Knowledge Retrieval}, a distinction not explicitly made in prior works \cite{rethink-demonstrations, function-vectors}.

\paragraph{Parametric Knowledge Retrieval} Parametric knowledge retrieval refers to questions that can be answered correctly by simply using the query and the knowledge within the model to perform a retrieval match (e.g., country-capital, country-currency). They were the focus in prior FV research \cite{which-fv-head,function-vectors}. We adapted 17 datasets from \citet{function-vectors}.
\paragraph{Contextual Knowledge Understanding} These tasks require understanding the content within a paragraph and using the information it provides to answer questions, such as hate speech classification and sentiment analysis.
The associations tend to be noiser than the parametric knowledge retrieval tasks.  We adapted 16 original datasets from \citet{rethink-demonstrations}. 

To guarantee a decent amount of testing samples, we keep 30\% of all samples for testing in the parametric knowledge retrieval datasets. For the contextual knowledge understanding datasets, we use their default train-test split available on Hugging Face. Detailed statistics of our datasets are listed in Table \ref{table:datasets} in Appendix \ref{appendix:datasets}.

\section{Behavioral Experiments}
We begin with behavioral experiments \cite{rethink-demonstrations} to test whether different architectures behave similarly in ICL.

\subsection{Label Randomization Experiments}
\paragraph{Method} We followed the same setup in \citet{rethink-demonstrations}: for each dataset, we sampled $k$ question-answer pairs as demonstrations with $k$ = 4, 8, 12, 16, and 32. To evaluate the performance, we found the logit indexes of the first token of each option, and then select the index with the maximum logits as the answer. We ran each experiment 5 times with 5 random seeds to measure the mean performance. For the parametric knowledge retrieval datasets without options, we sorted all possible answers to form the options \cite{function-vectors}. We conducted experiments under settings below:

\begin{enumerate}[noitemsep, topsep=0pt, left=1em]
    \item \textbf{No demo}: we give the model $k = 0$ demonstration, i.e. we present only the query.
    \item \textbf{$\alpha\%$ correct}: we augment our $k$ demonstrations such that $\alpha\%$ of the demonstrations are correct. We select $\alpha$ to be 0, 25, 50, 75.
    \item \textbf{Gold}: we give correct demonstrations to the model. This is equivalent to setting $\alpha = 100$.
    \item \textbf{Random}: the demonstrations' questions and answers are shuffled.
\end{enumerate}

\begin{figure}[t]
    \centering
    \includegraphics[width=\linewidth]{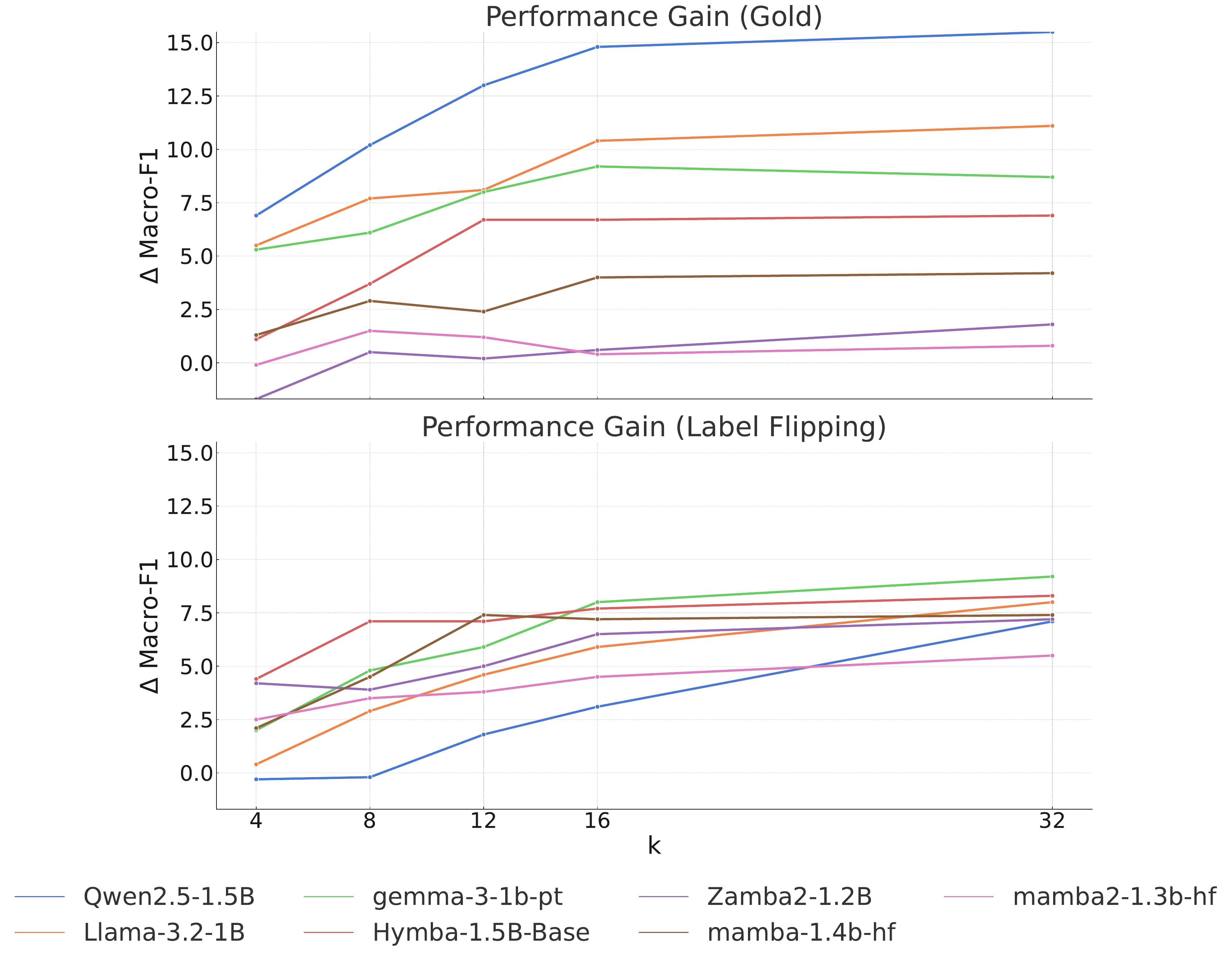}
    \caption{Performance gain of label flipping and gold conditions on contextual knowledge understanding datasets. Performance gain is the difference between the performance of gold or label flipping and the corresponding no demo conditions. Performance gain for other conditions are plotted in Figure \ref{fig:performance_gain_all} in Appendix \ref{appendix:additional-behavioral-results}.}
    \label{fig:performance_gain}
\end{figure}

\paragraph{Results} As presented in Figure \ref{fig:k=16}, results in parametric knowledge retrieval tasks demonstrate in-context learning being properly carried out. Surprisingly, we find that in contextual knowledge understanding datasets, while all models are affected by the percentage of correct demonstrations given, transformer-based models, \textsc{Hymba-1.5B-Base} and \textsc{mamba-1.4b-hf} are able to outperform \texttt{no demo} when gold demonstrations are provided, whereas \textsc{mamba2-1.3b-hf} and \textsc{Zamba2-1.2B} only demonstrate a marginal increment in performance. \textbf{Notice that the phenomenon that model performance varies significantly with different percentages of correct examples contradicts the findings of \citet{rethink-demonstrations}}. From the top figure of Figure \ref{fig:performance_gain}, we can observe that with more number of ICL examples, transformer models, \textsc{Hymba-1.5B-Base}, and \textsc{mamba-1.4b-hf} are able to perform better. However, this trend barely holds true for the models whose gold label performance merely outperforms \texttt{no demo}, namely \textsc{Zamba2-1.2B} and \textsc{mamba2-1.3b-hf}. \textbf{Under these regular settings, \textsc{Zamba2-1.2B} and \textsc{mamba2-1.3b-hf} have relatively weak ICL performance for tasks involving contextual knowledge understanding.}

\paragraph{Impact of $k$} \textbf{In general, the impact of $k$ shows consistent findings with \citet{rethink-demonstrations}, that is, models only benefit from more demonstrations when the majority of labels are reliable.} 
In the 0\% and 25\% correct settings, more examples actually hurt the performance. This is expected as more examples mislead the model and therefore cause lower performance. The 50\% correct and the random case yield similar curves as the 50\% correct case, as statistically they are almost equivalent. At 75\% correct, all models except \textsc{Zamba2-1.2B} and \textsc{mamba2-1.3b-hf} are demonstrating increments in performance with increased $k$, suggesting that they can guess the task from examples already. Performance plateaus mostly after $k = 8$ for all settings. Additional figures supporting these claims are available in Figures \ref{fig:k=4-classification} to \ref{fig:k=32-fvog} in Appendix \ref{appendix:additional-behavioral-results}.

\begin{figure*}[t]
    \centering
    \includegraphics[width=\linewidth]{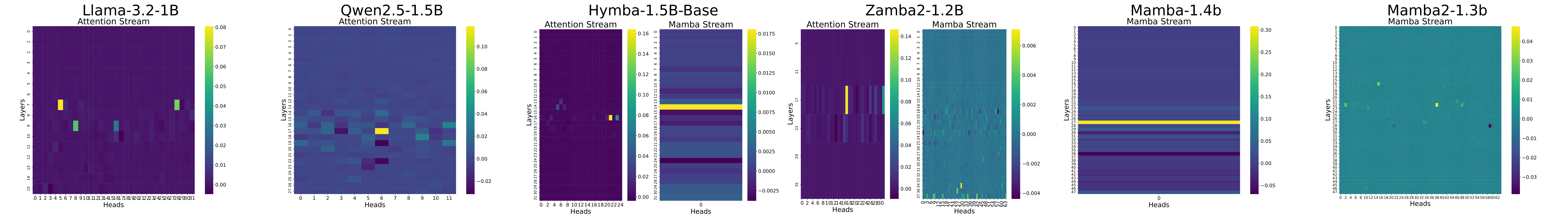}
    \caption{AIE heatmap for model heads on parametric knowledge retrieval datasets. [X-axis: head number; Y-axis: layer number.]}
    \label{fig:heatmap-fv_og}
\end{figure*}
\begin{figure*}[t]
    \centering
    \includegraphics[width=\linewidth]{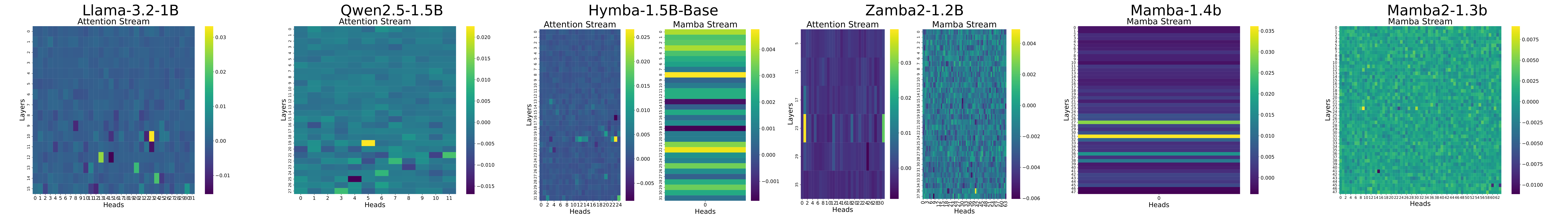}
    \caption{AIE heatmap for model heads on contextual knowledge understanding datasets. [X-axis: head number; Y-axis: layer number.]}
    \label{fig:heatmap-classification}
\end{figure*}
\begin{figure}
    \centering
    \includegraphics[width=\linewidth]{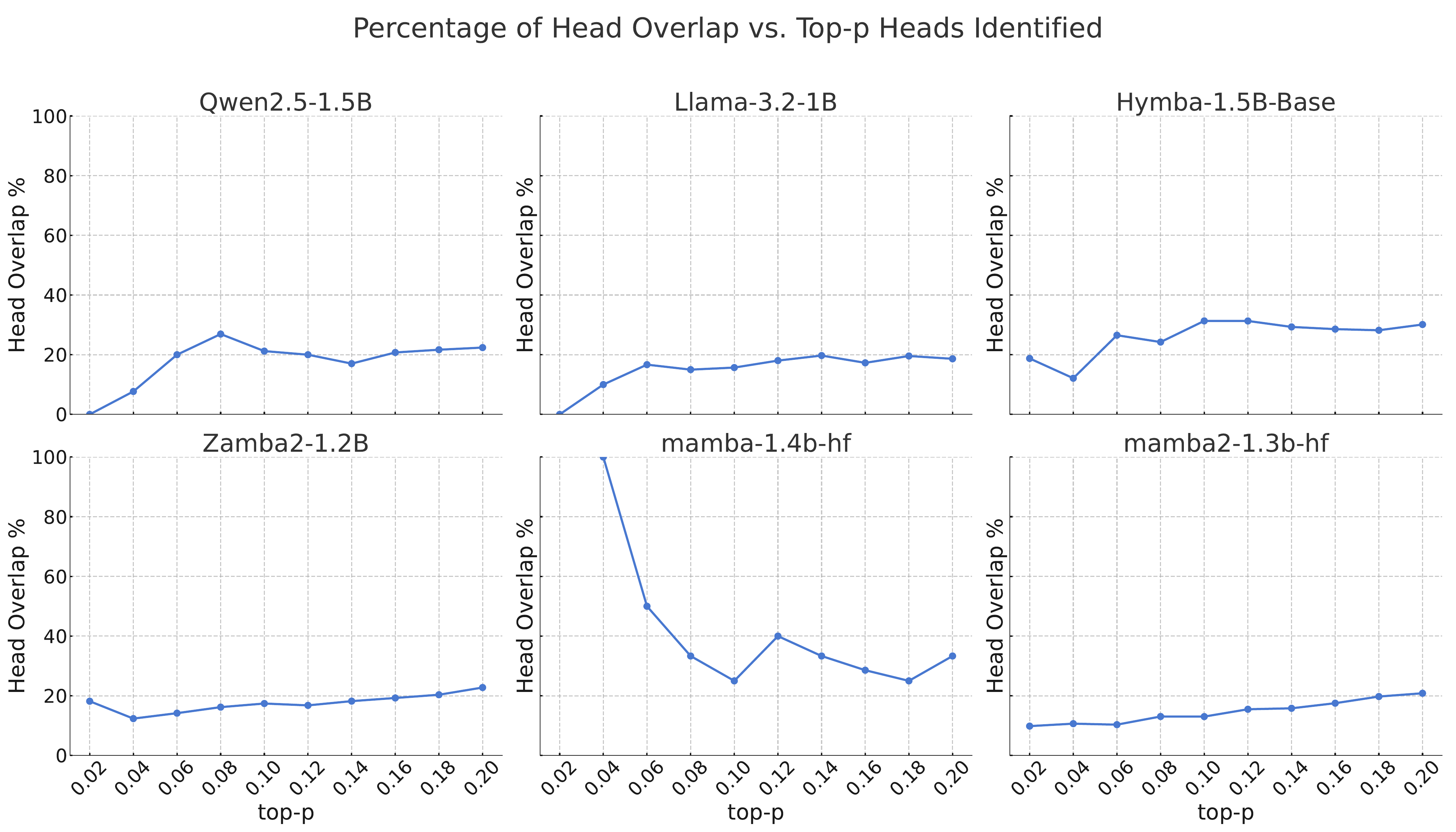}
    \caption{Percentage of intersection between top FV heads identified from contextual knowledge understanding datasets and parametric knowledge retrieval datasets. Entries are the percentage of heads in top-p that overlap. In general, different top heads are identified for different categories of tasks, as the overlaps are generally small. For \textsc{mamba-1.4b-hf}, the top-2\% heads yield an empty set.}
    \label{fig:head_overlap}
\end{figure}
\subsection{Label-flipping Experiments}
In the label randomization experiments above for contextual understanding and retrieval tasks, we have observed that \textsc{Zamba2-1.2B} and \textsc{mamba2-1.3B-hf} only demonstrate marginal gains on gold-label demonstrations. However, they experience a significant performance degradation when the given demonstrations are entirely wrong. We introduce the label-flipping experiments to look into this phenomenon.
\paragraph{Method}
To explore whether \textsc{mamba2-1.3b-hf} and \textsc{Zamba2-1.2B} can learn under a counterfactual setting and reduce the impact of memorization, we also introduced the label flipping setting: for each datapoint in the demonstrations and test set, we map the correct answer consistently to one of the incorrect options, and we expect the models to categorize queries to the flipped (incorrect) options. For example, for options \{hate, neutral, non-hate\},  we replaced ``hate`` with ``non-hate`` and used the flipped labels for evaluation. All mappings were strictly one-to-one to ensure that there are still systematic associations between questions and answers, but such associations should be absent in the training data. 

Notice that this is different from label randomization: label randomization randomizes the examplars' labels but expects the model to output the correct relationship, whereas label flipping expects the model to infer the new relationship from label-flipped examplars. The goal is to examine how well LLMs can pick up counterfactual relationships in context without the influence of domain knowledge. We do this for only the contextual knowledge understanding datasets.

\paragraph{Results}  The bottom panel of Figure \ref{fig:performance_gain} shows that all models are able to learn in-context the label-flipped tasks for contextual knowledge understanding tasks. Models whose performance gain used to be near zero or negative (\textsc{Zamba2-1.2B} and \textsc{mamba2-1.3b-hf}) are now above zero. We also see that these two models and \textsc{mamba-1.4b-hf} achieve more gain over the increasing number of ICL examples. Nevertheless, self-attention models are still able to achieve the biggest performance increment over the increasing number of ICL examples. \textbf{Evidence shows that all models, including \textsc{Zamba2-1.2B} and \textsc{mamba2-1.3b-hf}, are able to learn unseen associations in context, and that they all exhibit qualitatively the same learning curve.}

\begin{figure*}
    \centering
    \includegraphics[width=\linewidth]{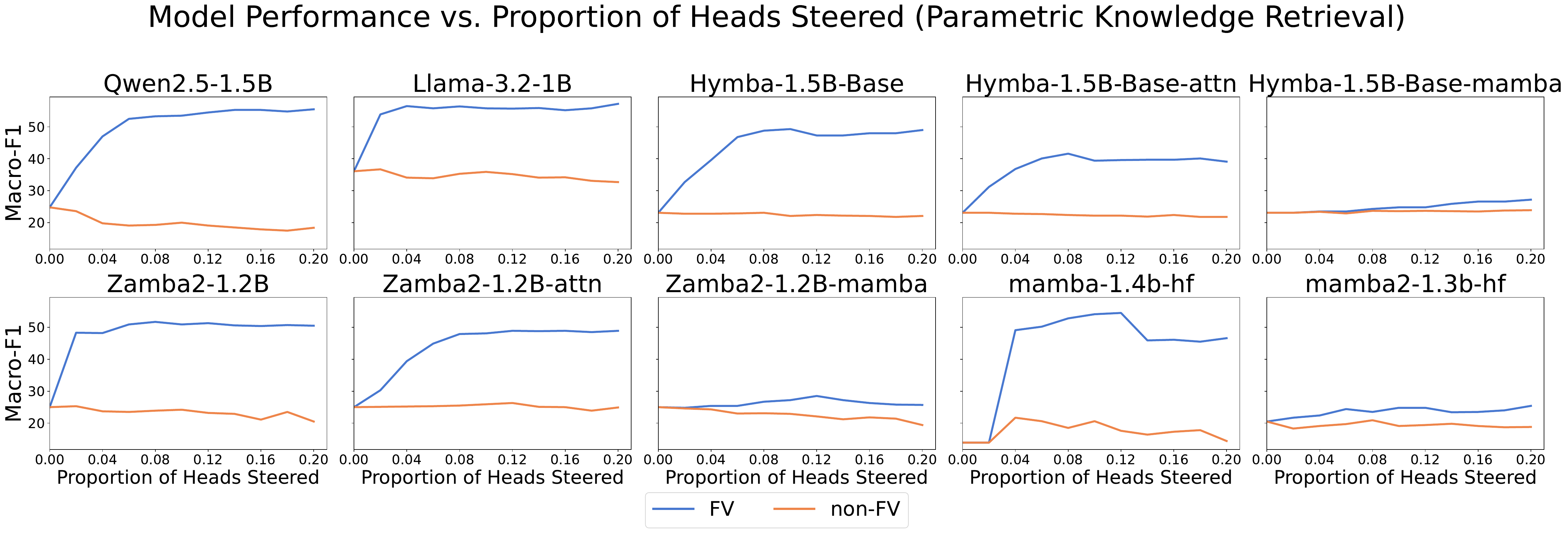}
    \caption{Steering results for models in parametric knowledge retrieval datasets. For hybrid models, we observe a significant performance increment for the self-attention stream while the mamba stream only demonstrates marginal improvement. Notably, \textsc{mamba2-1.3b-hf} does not seem to be significantly influenced by steering.}
    \label{fig:steer_mean_pool-fv_og}
\end{figure*}
\begin{figure*}[t]
    \centering
    \includegraphics[width=\linewidth]{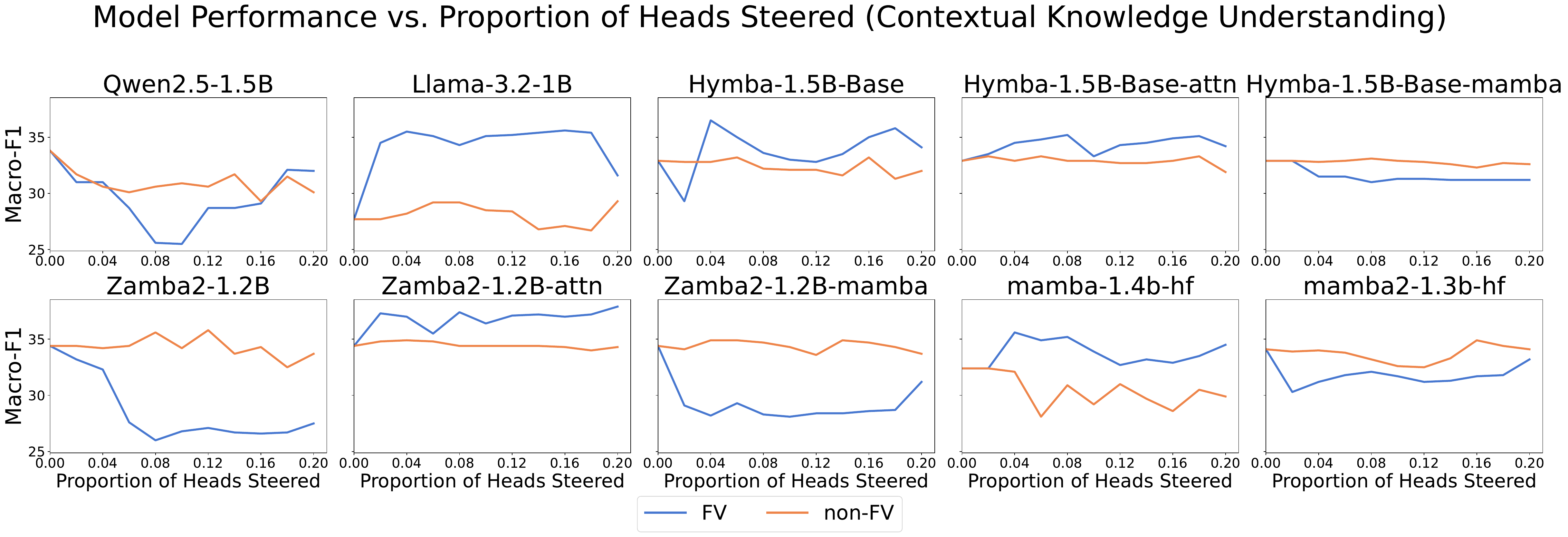}
    \caption{Steering results for models in contextual knowledge understanding datasets. Results demonstrate more fluctuations. It is worth noticing that steering the mamba stream in both hybrid models consistently decreases performance. Especially for \textsc{Zamba2-1.2B}, the decrease is so significant that it causes the overall performance to drop below non-FV steering, even though steering its self-attention stream still gives positive feedback. Surprisingly, steering \textsc{Qwen2.5-1.5B} hurts performance.}
    \label{fig:steer_mean_pool-classification}
\end{figure*}

\section{Mechanistic Interpretability Analysis}
The behavioral experiments have identified some differences in ICL across models. However, the internal mechanism behind this phenomenon remains unclear.
In this section, we will examine the FV heads in \textsc{Qwen2.5-1.5B}, \textsc{Llama-3.2-1B}, \textsc{Hymba-1.5B-Base}, \textsc{Zamba2-1.2B}, \textsc{mamba-1.4b-hf}, and \textsc{mamba2-1.3b-hf}.
\begin{figure*}[t]
    \centering
    \includegraphics[width=\linewidth]{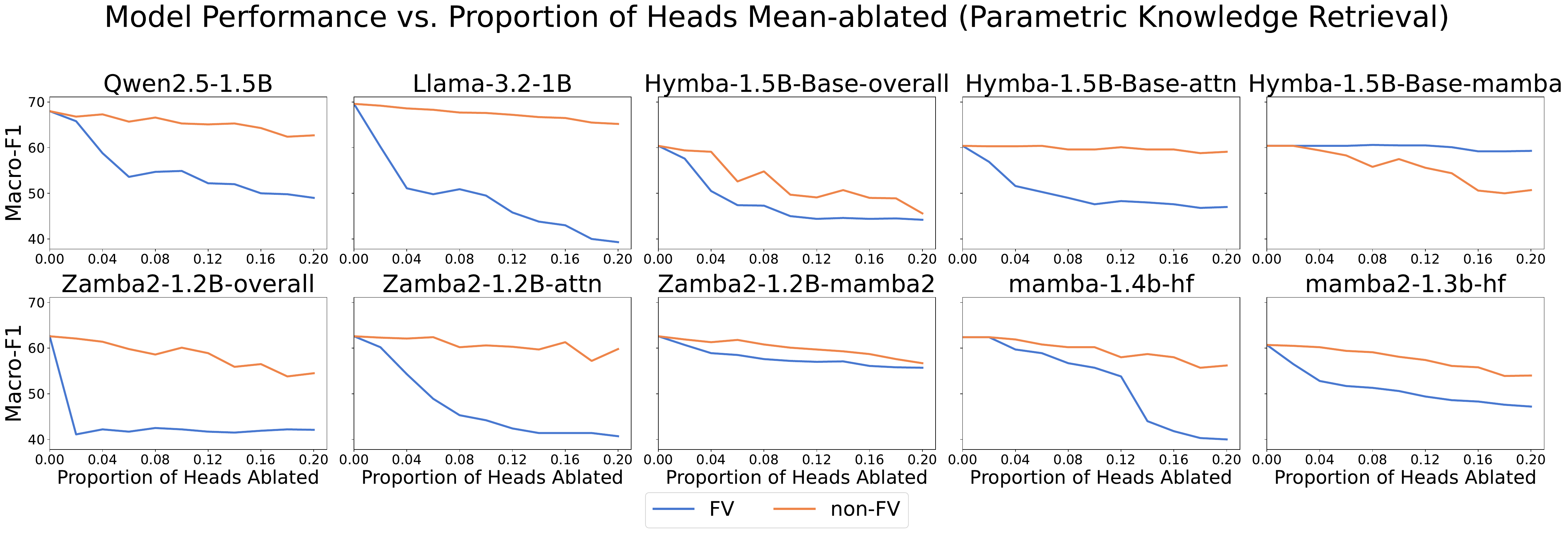}
    \caption{Mean ablation results for parametric knowledge retrieval ICL.}
    \label{fig:mean_ablation-fv_og}
\end{figure*}
\begin{figure*}
    \centering
    \includegraphics[width=\linewidth]{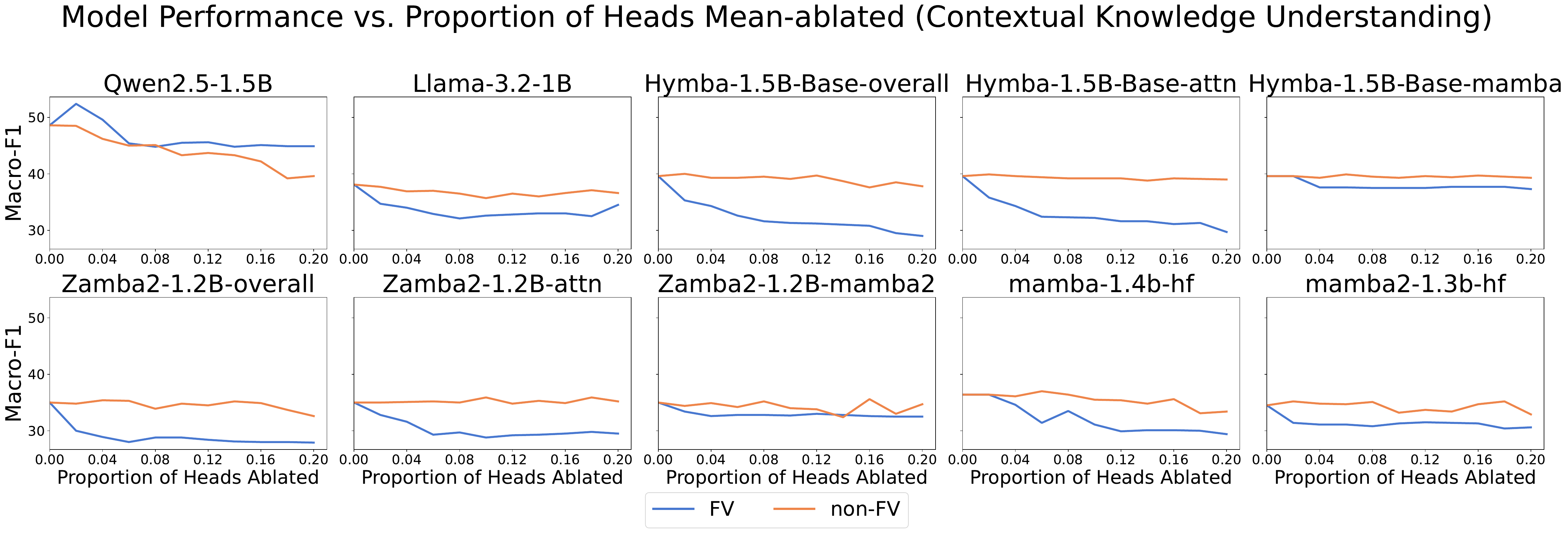}
    \caption{Mean ablation results for contextual knowledge understanding ICL.}
    \label{fig:mean_ablation-classification}
\end{figure*}

\subsection{Identifying FV Heads}
We replicated the method of \citet{function-vectors} on our models to identify FV heads. Originally, this was done only for transformers. For SSMs and hybrid models, we treated the SSM heads as analogous to attention heads. 

We first extracted the FVs for all heads at all layers from all streams at the very last token of the input prompt. Since Mamba and the Mamba stream of \textsc{Hymba-1.5B-Base} was not multi-head, we treated it as one head. Mathematically put, let $p_i^t \in P_t$ be a prompt in dataset $P_t$ representing task $t$, for each attention head $a_{lj}$ at layer $l$, we wish to take the task-dependent output value of this attention head, calculated as:
$$\bar{a_{lj}}^t = \frac{1}{|P_t|} \sum_{p_i^t \in P_t} a_{lj}(p_i^t)$$
We used the $k = 10$ randomly chosen demonstrations, and we extracted the FVs on the validation set. After that, for each task, we calculated the average logit-level percentage shift towards the correct logit as we replaced each head with its corresponding $\bar{a_{lj}}^t$, namely average indirect effect (AIE), on the random demonstrations setting. Formally:
$$\text{AIE}(a_{lj}) = \frac{1}{|\mathcal{T}|} \sum_{t \in \mathcal{T}} \frac{1}{\tilde{P_t}} \sum_{\tilde{p}_i^t \in \tilde{P_t}} \text{CIE}(a_{lj} | \tilde{p}_i^t)$$
where $\tilde{p}_i^t \in \tilde{P_t}$ is the corrupted prompt with random demonstrations from the random demonstration dataset $\tilde{P_t}$, $t \in \mathcal{T}$ is a dataset in task $\mathcal{T}$, and CIE is the causal indirect effect defined as:
$$\text{CIE}(a_{lj} | \tilde{p}_i^t) = f(\tilde{p}_i^t | a_{lj} := \bar{a_{lj}}^t)[y_{iq}] - f(\tilde{p}_i^t)[y_{iq}]$$
where $f$ is the model's forward pass, and $y_{iq}$ is the token of the correct answer.

We reserved 100 random samples from the training set to compute AIE, and the ICL examples are sampled at random from the remaining training samples. Since AIE is compute-heavy, we used $k = 10$ ICL examples over 25 random samples from the development split over only one random seed. We then evaluated the steering performance on the test set. 

\subsection{Locations of FV heads}
\textbf{Top FV heads are highly consistent and concentrated for parametric knowledge retrieval ICL but not contextual understanding ICL.} Figures \ref{fig:heatmap-fv_og-hymba} and \ref{fig:heatmap-classification-hymba} show that, in \textsc{Hymba-1.5B-Base}, certain heads are always activated in the parametric knowledge retrieval ICL, but the activations are less concentrated in contextual knowledge ICL. Such patterns also exist in all other models with full heatmaps in Appendix \ref{appendix:additional-interpretability-results}. As shown in Figures \ref{fig:heatmap-fv_og} and \ref{fig:heatmap-classification}, under different tasks, all models except \textsc{mamba-1.4b-hf} activate different heads for the two different categories of tasks. We quantified the percentage of overlapping top-p FV heads for the two categories in Figure \ref{fig:head_overlap}. \textbf{Inspection suggests that the top FVs for the two kinds of ICL tasks do not necessarily overlap in most models except for \textsc{mamba-1.4b-hf}.}

\subsection{Function vector internevtion}
In the last section, we had identified the top FVs. To causally validate their roles, we conducted two intervention experiments.

\begin{figure*}[t]
    \centering
    \includegraphics[width=\linewidth]{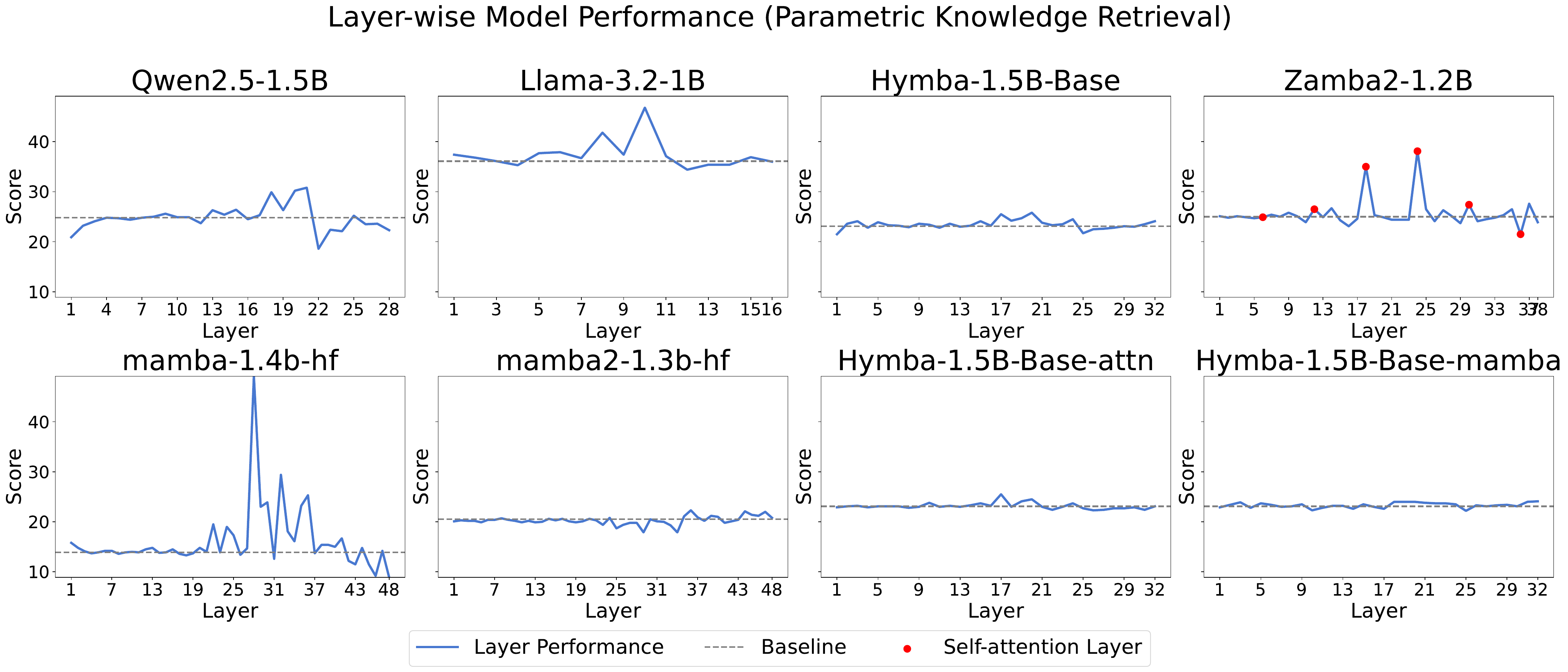}
    \caption{Layer-wise steering results for models in parametric knowledge retrieval datasets. The self-attention stream remains dominant in both hybrid models. We also see \textsc{mamba-1.4b-hf}'s middle layers playing a significant role.}
    \label{fig:layer_steer-fv_og}
\end{figure*}
\begin{figure*}
    \centering
    \includegraphics[width=\linewidth]{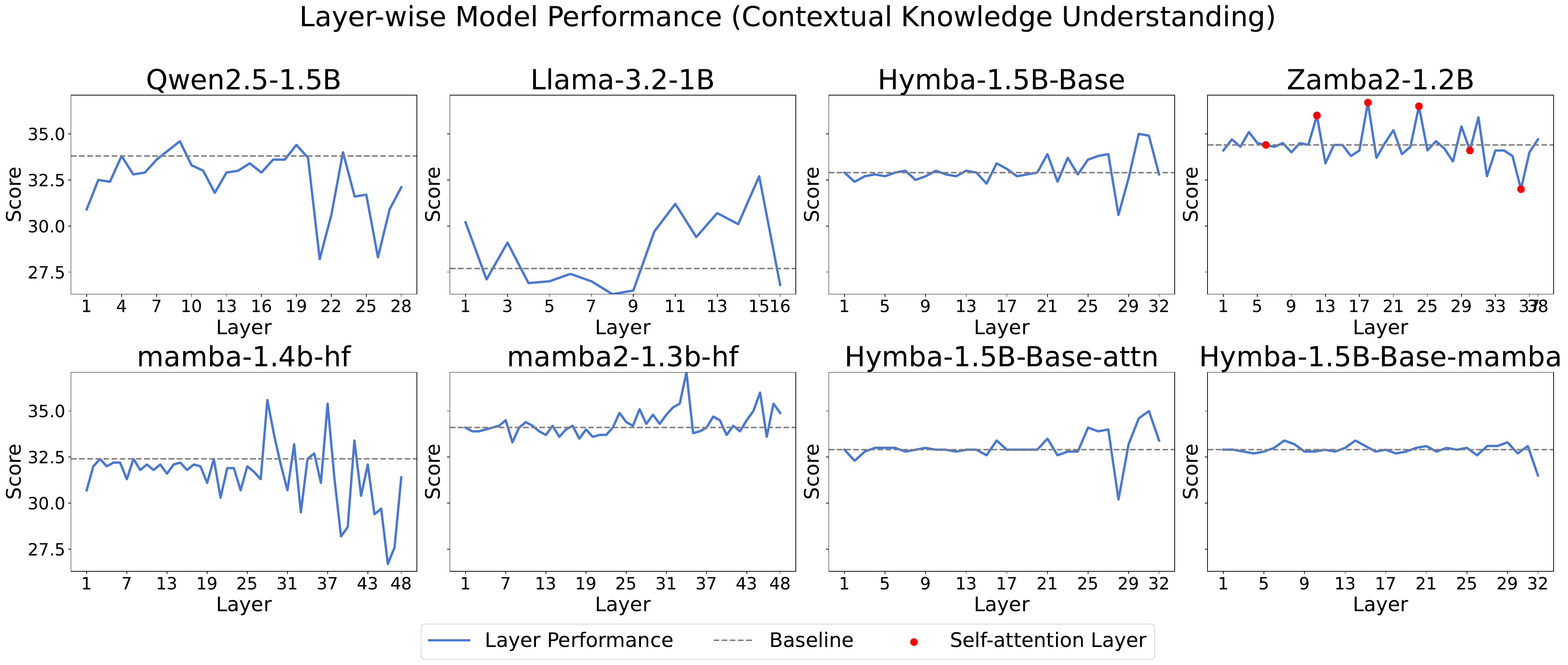}
    \caption{Layer-wise steering results for models in contextual knowledge understanding datasets. The self-attention stream remains dominant in both hybrid models.}
    \label{fig:layer_steer-classification}
\end{figure*}

\paragraph{Steering Function Vector Heads} 
We took the no demo variants of each dataset, then added $\bar{a_{lj}}^t$ to the output of that head at the last token position. Formally put, let $a_{lj}^t$ be the value of the last token of output from attention head $j$ in layer $l$, we add $\bar{a_{lj}}^t$ calculated before to $a_{lj}^t$:

$${a_{lj}^t}' = a_{lj}^t + \bar{a_{lj}}^t$$

We then evaluated the performance of models. We did this for the top 2 to 20 percent of heads, selected by AIE. For hybrid models, we first selected the top heads on both streams, then did the same on each stream individually, to see the effectiveness of both streams. As a comparison, we first steered 2 to 20 percent of random heads outside of the top 20\% FV heads (non-FV heads) identified by AIE. We also steered all heads at each layer only.

\paragraph{Ablating function vector heads} Adopting the experiment in \citet{which-fv-head}, we replaced the last token of the outputs of selected FV heads with zeros or the mean of this head's activation at the last token position over all datasets within its category. We did this for the top 2 to 20 percent of heads, selected by AIE. We did the same as the steering experiment above for hybrid models. We also ablated random 2 to 20 percent of heads outside of the top 20\% identified FV heads (non-FV heads) as a comparison. We would focus on mean ablations as in \citet{icl-heads}, because zero ablations may cause the out-of-distribution problem \cite{OOD1, OOD2, OOD3}. 

\subsection{Findings}
Our two intervention-based analyses yield highly consistent results, and the main findings are summarized below. 

\textbf{FV heads drive ICL in self-attention models and Mamba, but not so much in Mamba2.} We observe in Figures \ref{fig:steer_mean_pool-fv_og} and \ref{fig:steer_mean_pool-classification} that the steering performance increment of \textsc{mamba2-1.3b-hf} is insignificant compared to other models. However, unlike in the steering experiments, Figures \ref{fig:mean_ablation-fv_og} and \ref{fig:mean_ablation-classification} show that \textsc{mamba2-1.3b-hf}'s performance change in ablations is comparable with other models in both categories. Along with our steering results, we can claim that \textsc{mamba2-1.3b-hf} does not depend on FV heads to retrieve task knowledge, but rather uses its heads for other purposes. We hypothesize that since \textsc{mamba2-1.4b-hf} is multi-head, each head has a smaller number of hidden dimensions and therefore is less capable of capturing task-specific information at the head level.

\textbf{FV heads can account for the ICL performance in parametric knowledge retrieval tasks, but not in contextual knowledge understanding tasks.} In Figures \ref{fig:steer_mean_pool-fv_og} and \ref{fig:steer_mean_pool-classification}, we can observe that steering FV heads are mostly effective in the parametric knowledge retrieval dataset, whereas there are significantly more fluctuations in the contextual knowledge understanding datasets. Performance increment for steering contextual knowledge understanding datasets is significantly less than steering parametric data retrieval datasets. Notably, we see a consistent decrease in performance when steering the Mamba streams of both hybrid models in the contextual knowledge understanding datasets. It is likely that the Mamba stream serves other functionalities than the FV heads in the attention stream. With these observations, we can conclude that FVs are not key in contextual knowledge understanding datasets.

\textbf{For hybrid models, their ICL capabilities are primarily controlled by FV heads at self-attention layers.} To further study FV heads in these models, we will focus on parametric knowledge retrieval datasets. Looking at the figures of \textsc{Hymba-1.5B-Base} and \textsc{Zamba2-1.2B}, FV heads are primarily located in the attention stream of hybrid models. Nevertheless, the Mamba stream of \textsc{Hymba-1.5B-Base} is also important, as simply steering the attention stream cannot match the performance of steering both streams. This can be further justified by the results in Appendix \ref{appendix:additional-hymba-results}. We also find the non-FV heads identified in \textsc{Hymba-1.5B-Base} are very important, as ablating them causes more performance drop than the FV heads in the Mamba stream (Figure \ref{fig:mean_ablation-fv_og}). This is likely caused by \textsc{Hymba-1.5B-Base} taking the mean of attention and Mamba stream output at each layer, so at some layers where FV heads are located but the corresponding Mamba head is not identified, this may cause a problem.

\textbf{Steering FVs in the middle or later layers tends to improve ICL performance.} Layerwise steering results in Figures \ref{fig:layer_steer-fv_og} and \ref{fig:layer_steer-classification} also reinforce our findings. We can easily identify that the self-attention layers in \textsc{Zamba2-1.2B}, especially the middle few, play a significant role in both sets of datasets. We also identify that the overall performance of \textsc{Hymba-1.5B-Base} is mostly governed by its attention stream. It is worth noticing that transformer models in parametric knowledge ICL react quite significantly to layer-wise steering. However, the layers that yield the highest performance increments do not necessarily contain the top identified FV heads, suggesting that they are more sensitive to layer-wise steering.

\section{Conclusions}
In this research, we performed an in-depth analysis of ICL on several mainstream architectures. Through a combination of behavioral and mechanistic analysis,  we report several new findings concerning how FVs contribute to ICL in different architectures and task settings. Our work extends and refines the understanding of FVs from transformers to SSMs and hybrid models.

\section{Limitations}
We acknowledge that our study is still limited in several ways. ICL is a complex capability in LLMs that might involve multiple mechanisms in synchrony. In this study, we only focus on one potential mechanism, namely, FV heads. Yet there might be other mechanisms, including induction heads. Future studies should investigate more mechanisms to uncover the intricacies of ICL.

Due to limited computing budget, our analyses are limited to pretrained LLMs, like all prior works \cite{function-vectors,icl-heads,rethink-demonstrations}. While we have made our best attempts to control for the experiment settings and model parameters, we have no control over the pretraining materials. We speculate that pretraining materials and procedures could also impact ICL capabilities and the locations of FV heads, since both \textsc{Qwen2.5-1.5B} and \textsc{Llama3.2-1B} still exhibit slightly different internals despite using similar self-attention layers. More well-designed experiments that strictly control for training data, hyperparameters, and training procedures across architectures can better clarify these issues. We will leave these to future studies.

\section{Ethics statement}
Our research primarily makes empirical contributions toward the internal mechanisms of LLMs. While most findings do not have direct practical applications, a better understanding of the internal mechanisms may be exploited by malicious users to actively jailbreak or take control over deployed LLMs, potentially leading to undesirable societal risks. 

\section{Acknowledgments}
We thank three anonymous reviewers and the area chairs for their thoughtful comments on the original manuscript. This research was enabled in part through the computational resources provided by Advanced Research Computing at the University of British Columbia and the Digital Research Alliance of Canada. The research activities were also supported by the NSERC Discovery Grant and the CFI-JELF Grant awarded to JZ.
\bibliography{custom}

\appendix

\section{Datasets}
\label{appendix:datasets}
\begin{table}[H]
    \small
    \begin{tabular}{p{5cm}ll}
        \toprule
        Dataset                           & \# Train     & \# Eval    \\
        \midrule
        \multicolumn{3}{l}{\textit{Datasets category: contextual knowledge understanding}}         \\
        financial\_phrasebank \cite{financial_phrasebank} & 1811          & 453           \\
        poem\_sentiment \cite{poem_sentiment} & 892           & 105           \\
        medical\_questions\_pairs \cite{medical_question_pairs} & 2438          & 610           \\
        glue-mrpc \cite{glue-mrpc} & 3668          & 408           \\
        glue-wnli \cite{glue-wnli} & 635           & 71            \\
        climate\_fever \cite{climate_fever} & 1228          & 307           \\
        glue-rte \cite{glue-rte1, glue-rte2, glue-rte3}  & 2490          & 277           \\
        superglue-cb \cite{superglue-cb} & 250           & 56            \\
        sick \cite{sick} & 4439          & 495           \\
        hate\_speech18 \cite{hate_speech18} & 8562          & 2141          \\
        ethos-national\_origin \cite{ethos} & 346           & 87            \\
        ethos-race \cite{ethos} & 346           & 87            \\
        ethos-religion \cite{ethos} & 346           & 87            \\
        tweet\_eval-hate \cite{tweet_eval-hate} & 8993          & 999           \\
        tweet\_eval-stance\_athesim \cite{tweet_eval-stance} & 461           & 52            \\
        tweet\_eval-stance\_feminist \cite{tweet_eval-stance} & 597           & 67            \\
        \midrule
        \multicolumn{3}{l}{\textit{Datasets category: parametric knowledge retrieval}}         \\
        antonym \cite{antonym-synonym} & 1678          & 720           \\
        capitalize\_first\_letter \cite{function-vectors} & 569           & 244           \\
        capitalize \cite{function-vectors} & 569           & 244           \\
        country-capital \cite{function-vectors} & 137           & 60            \\
        country-currency \cite{function-vectors} & 137           & 60            \\
        english-french \cite{translation} & 3288          & 1410          \\
        english-german \cite{translation} & 3288          & 1410          \\
        english-spanish \cite{translation} & 3639          & 1560          \\
        landmark-country \cite{linearity-decoding} & 585           & 251           \\
        lowercase\_first\_letter \cite{function-vectors} & 569           & 245           \\
        national\_parks \cite{function-vectors} & 315           & 136           \\
        park-country \cite{function-vectors} & 524           & 225           \\
        person-sport \cite{linearity-decoding} & 222           & 96            \\
        present-past \cite{function-vectors} & 205           & 88            \\
        product-company \cite{linearity-decoding} & 365           & 157           \\
        singular-plural \cite{function-vectors} & 143           & 62            \\
        synonym \cite{antonym-synonym} & 2015          & 865           \\
        \bottomrule
    \end{tabular}
    \caption{All datasets being used. In experiments, we randomly sample $k$ samples from \# Train.}
    \label{table:datasets}
\end{table}

\section{Additional Behavioral Experiment Results}
\label{appendix:additional-behavioral-results}
\begin{figure*}
    \centering
    \includegraphics[width=0.8\linewidth]{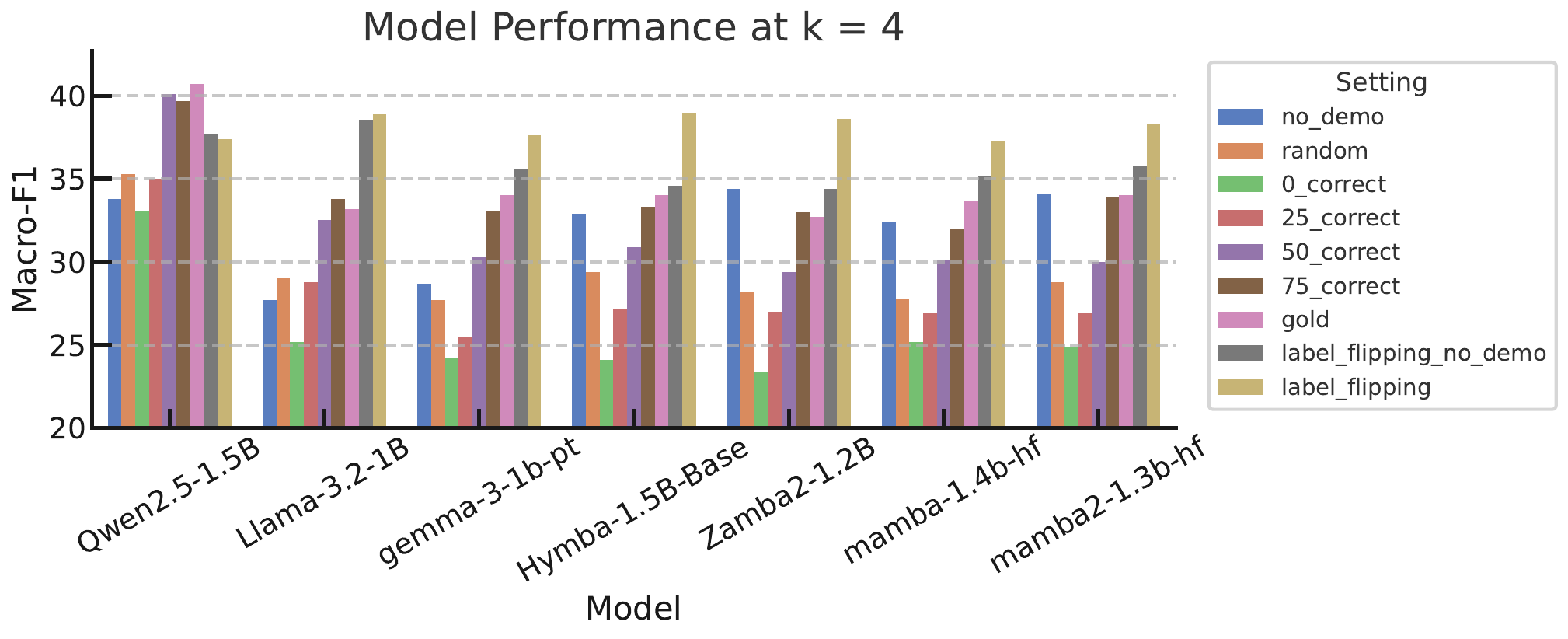}
    \caption{$k = 4$ results on contextual knowledge understanding datasets, including label flipping experiments.}
    \label{fig:k=4-classification}
\end{figure*}
\begin{figure*}
    \centering
    \includegraphics[width=0.8\linewidth]{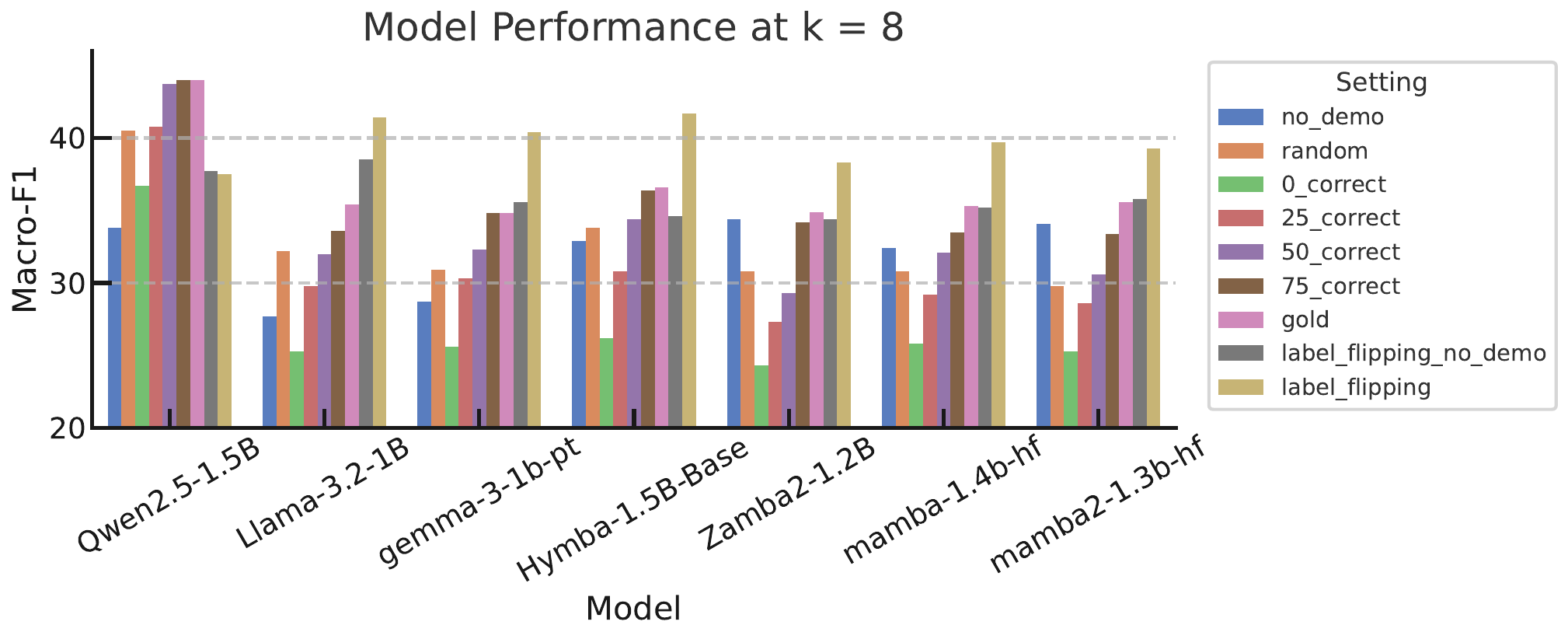}
    \caption{$k = 8$ results on contextual knowledge understanding datasets, including label flipping experiments.}
    \label{fig:k=8-classification}
\end{figure*}
\begin{figure*}
    \centering
    \includegraphics[width=0.8\linewidth]{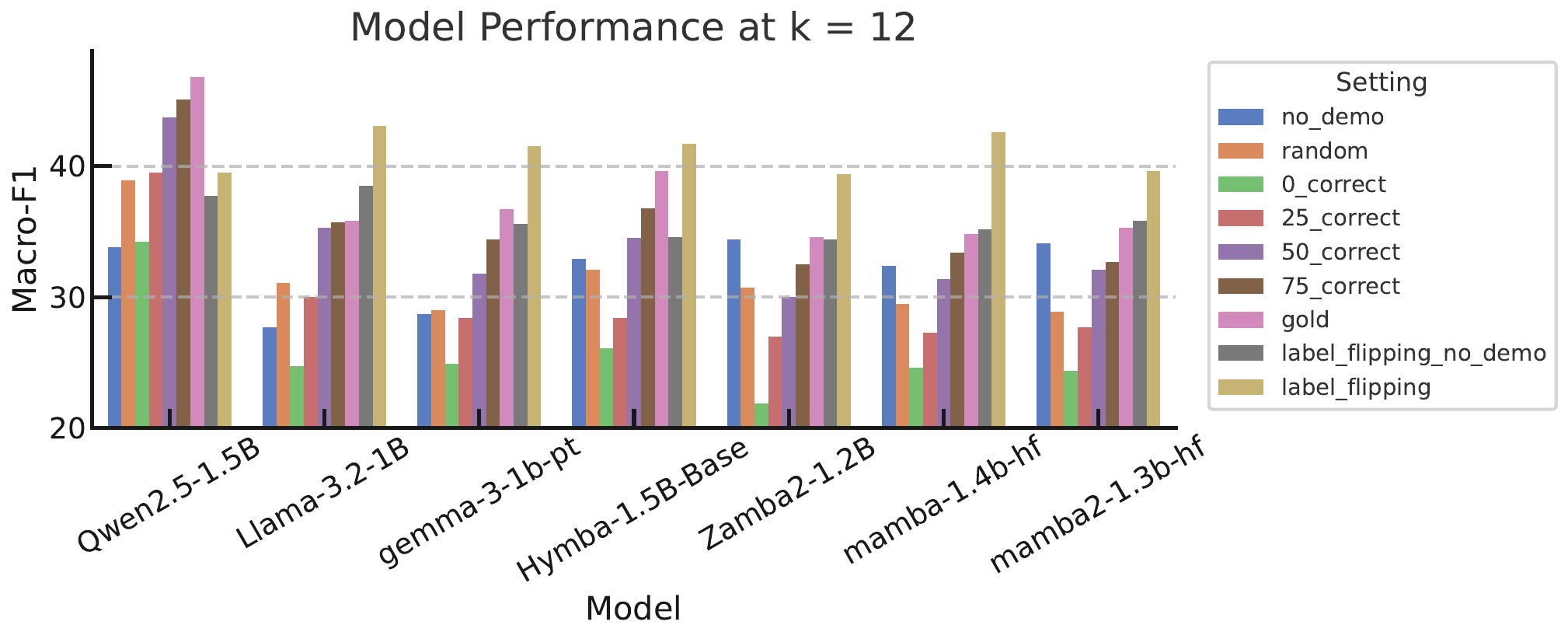}
    \caption{$k = 12$ results on contextual knowledge understanding datasets, including label flipping experiments.}
    \label{fig:k=12-classification}
\end{figure*}
\begin{figure*}[t]
    \centering
    \includegraphics[width=0.8\linewidth]{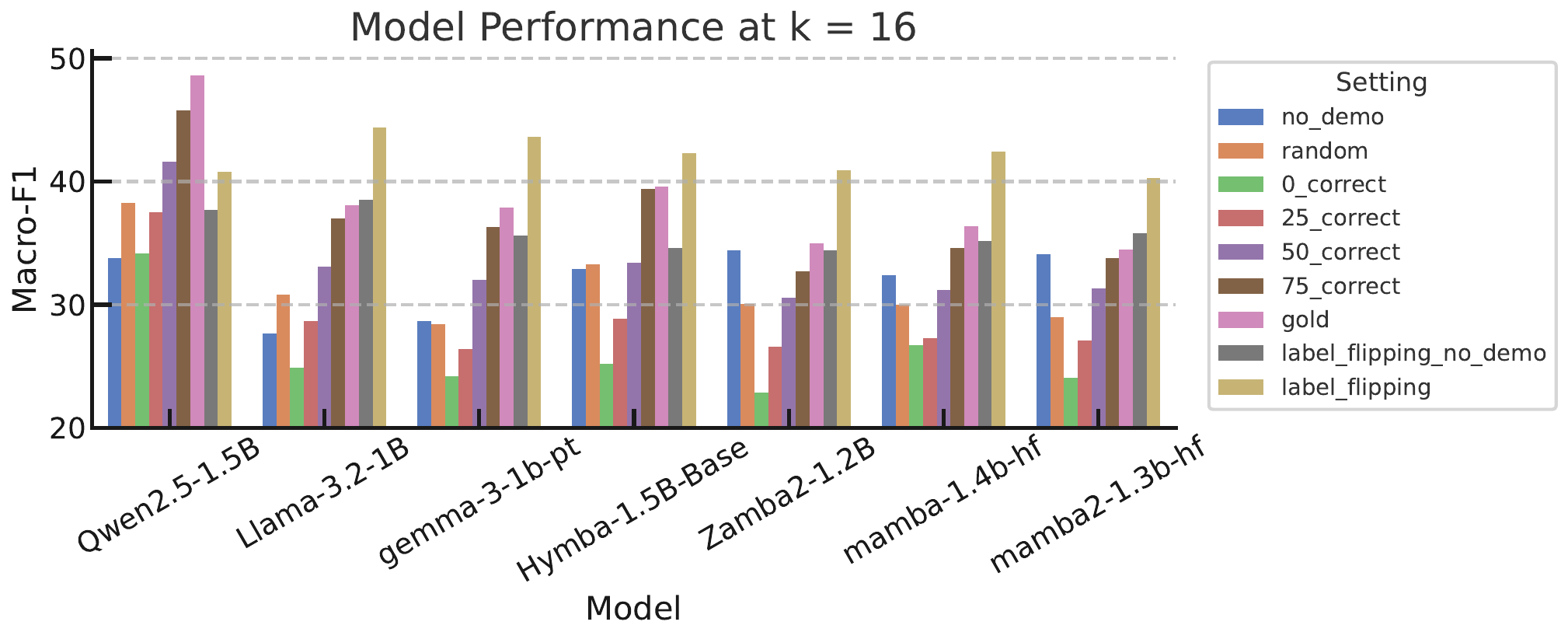}
    \caption{$k = 16$ results on contextual knowledge understanding datasets, including label flipping experiments.}
    \label{fig:k=16-classification}
\end{figure*}
\begin{figure*}
    \centering
    \includegraphics[width=0.8\linewidth]{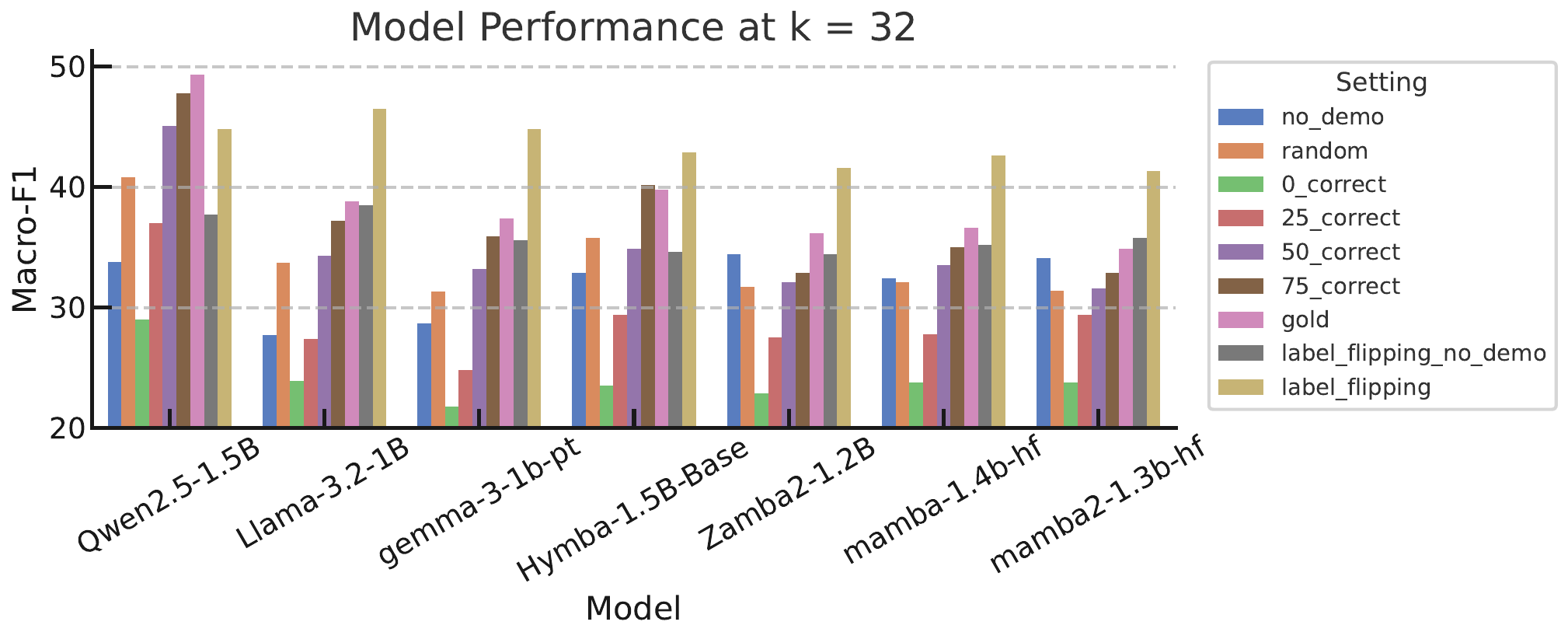}
    \caption{$k = 32$ results on contextual knowledge understanding datasets, including label flipping experiments.}
    \label{fig:k=32-classification}
\end{figure*}
\begin{figure*}
    \centering
    \includegraphics[width=0.8\linewidth]{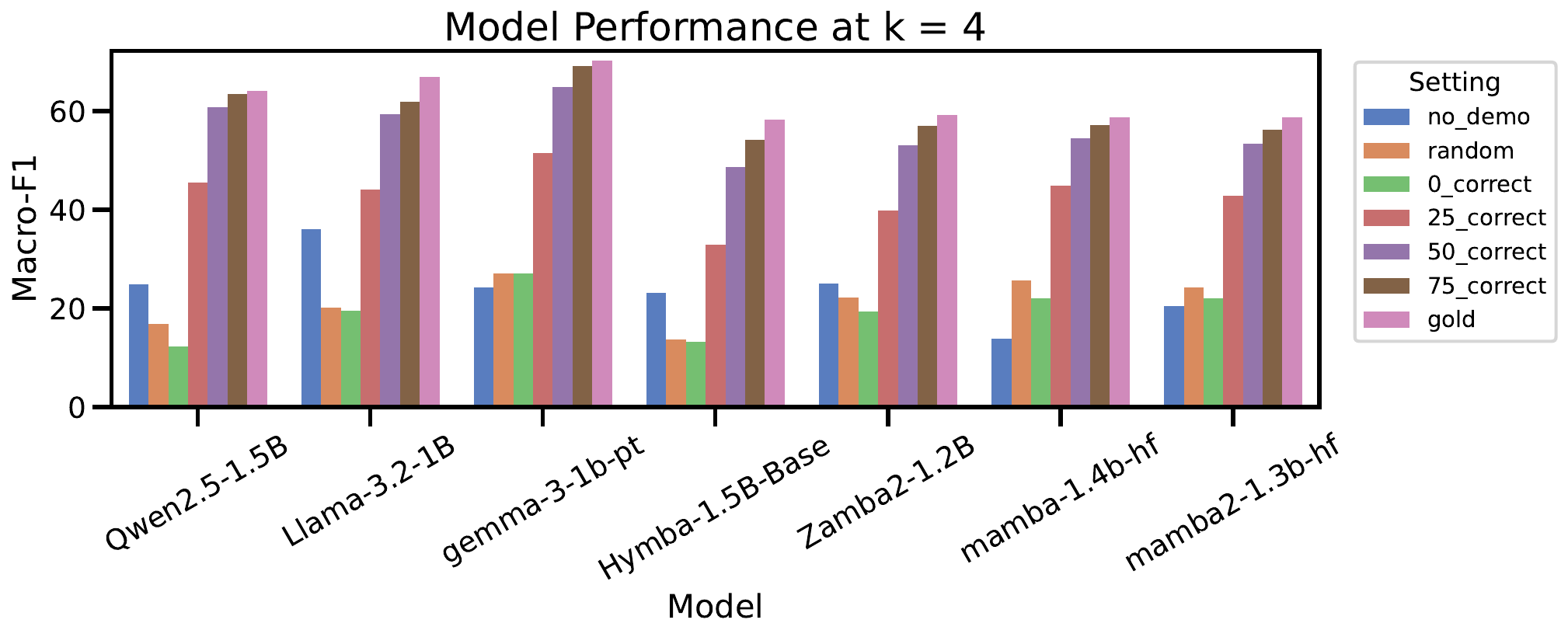}
    \caption{$k = 4$ results on parametric knowledge retrieval datasets, on initial setting.}
    \label{fig:k=4-fvog}
\end{figure*}
\begin{figure*}
    \centering
    \includegraphics[width=0.8\linewidth]{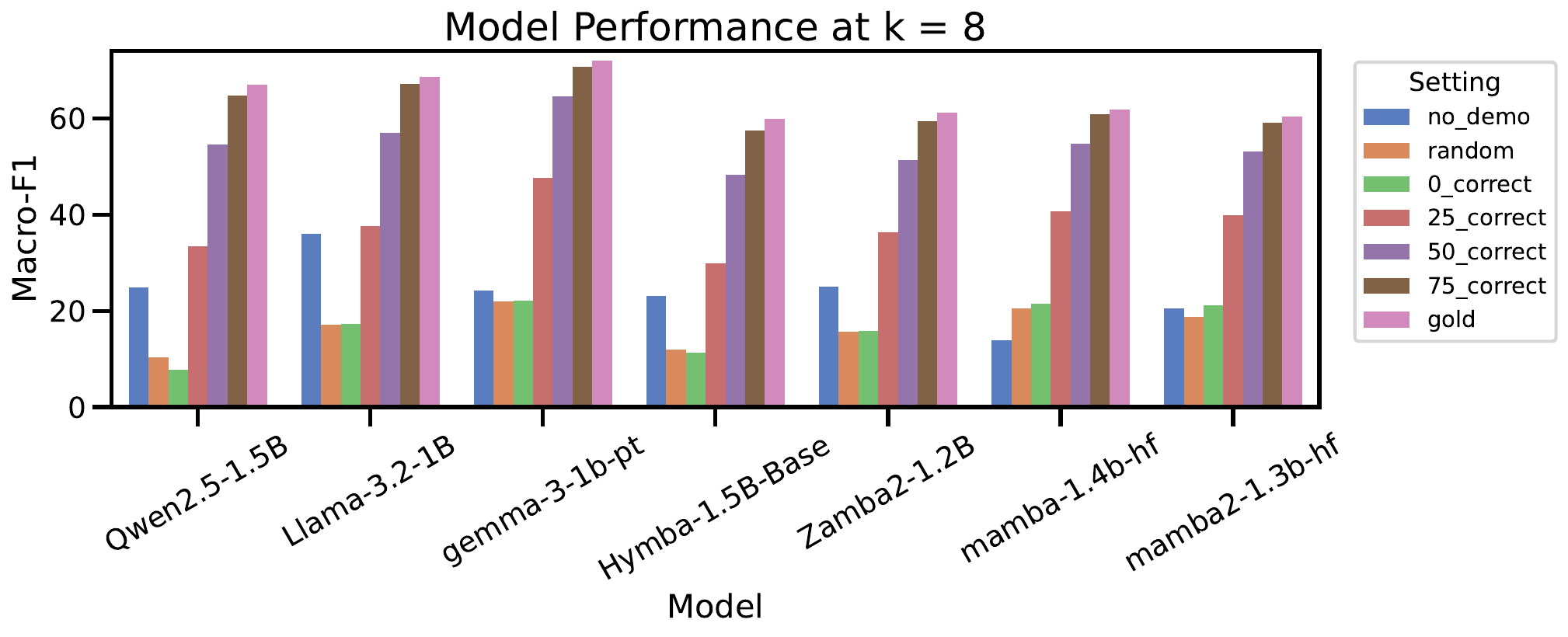}
    \caption{$k = 8$ results on parametric knowledge retrieval datasets, on initial setting.}
    \label{fig:k=8-fvog}
\end{figure*}
\begin{figure*}
    \centering
    \includegraphics[width=0.8\linewidth]{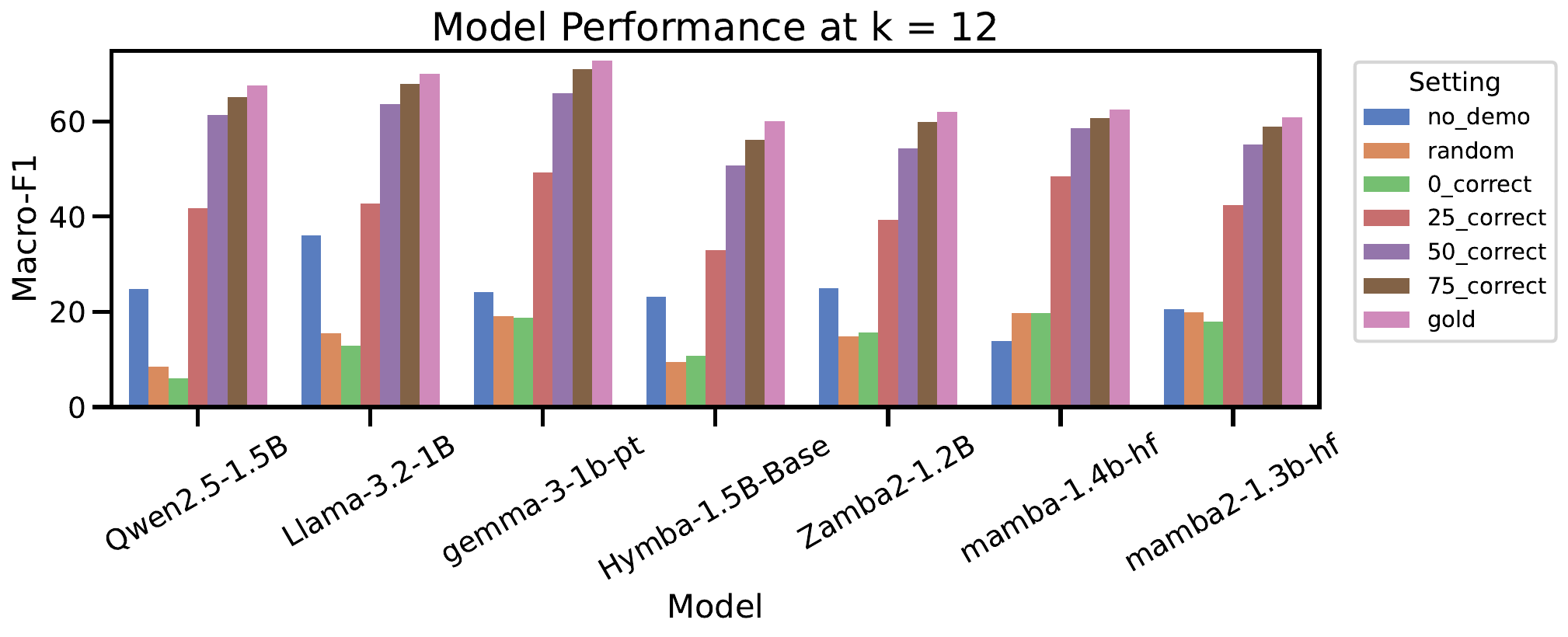}
    \caption{$k = 12$ results on parametric knowledge retrieval datasets, on initial setting.}
    \label{fig:k=12-fvog}
\end{figure*}
\begin{figure*}
    \centering
    \includegraphics[width=0.8\linewidth]{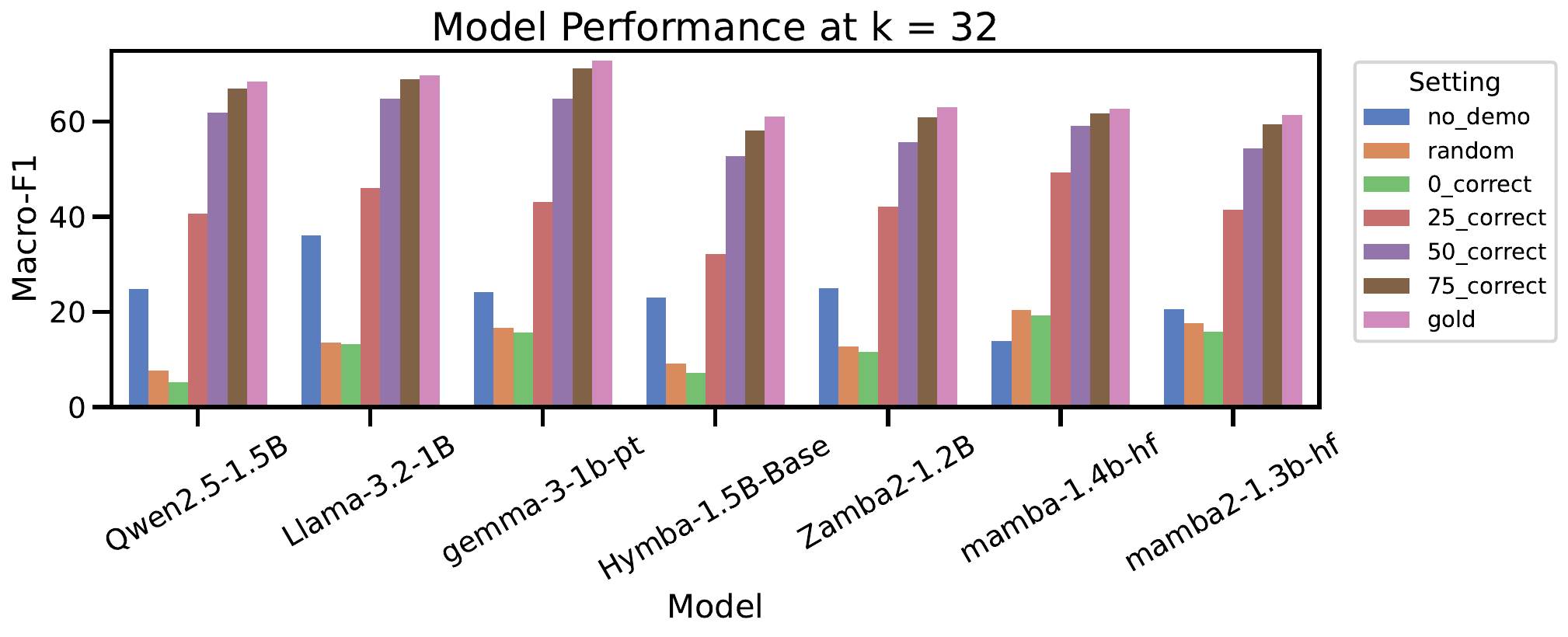}
    \caption{$k = 32$ results on parametric knowledge retrieval datasets, on initial setting.}
    \label{fig:k=32-fvog}
\end{figure*}
\begin{figure*}
    \centering
    \includegraphics[width=0.8\linewidth]{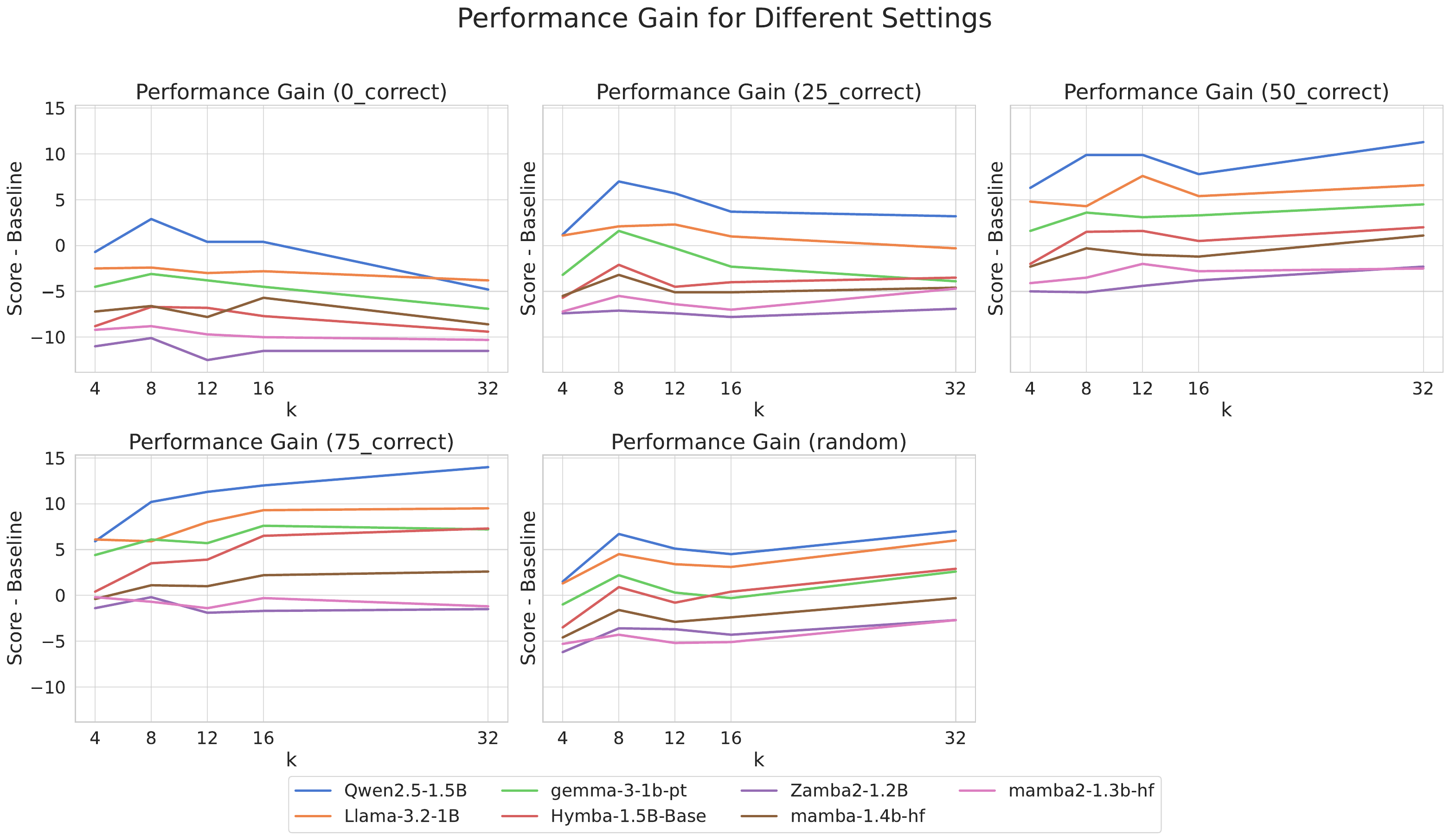}
    \caption{Performance gain of other conditions on contextual knowledge understanding datasets.}
    \label{fig:performance_gain_all}
\end{figure*}
\begin{table*}
    \centering
    \resizebox{\textwidth}{!}{
        \begin{tabular}{lrrrrrrr}
        \toprule
        \textbf{setting}                      & \textbf{Qwen2.5-1.5B} & \textbf{Llama-3.2-1B} & \textbf{gemma-3-1b-pt} & \textbf{\textsc{Hymba-1.5B-Base}} & \textbf{\textsc{Zamba2-1.2B}} & \textbf{mamba-1.4b-hf} & \textbf{mamba2-1.3b-hf} \\ \midrule
        no\_demo                     & 33.8         & 27.7         & 28.7          & 32.9            & 34.5        & 34.6          & 34.1           \\
        incorrect\_mapping\_no\_demo & 37.7         & 38.5         & 35.6          & 34.6            & 34.4        & 35.2          & 35.8           \\ \midrule
        \textit{k = 4} \\
        0\_correct                   & 33.1         & 25.2         & 24.2          & 24.1            & 23.4        & 25.2          & 24.9           \\
        25\_correct                  & 35.0         & 28.8         & 25.5          & 27.2            & 27.0        & 26.9          & 26.9           \\
        50\_correct                  & 40.1         & 32.5         & 30.3          & 30.9            & 29.4        & 30.1          & 30.0           \\
        75\_correct                  & 39.7         & 33.8         & 33.1          & 33.3            & 33.0        & 32.0          & 33.9           \\
        gold                         & 40.7         & 33.2         & 34.0          & 34.0            & 32.7        & 33.7          & 34.0           \\
        random                       & 35.3         & 29.0         & 27.7          & 29.4            & 28.2        & 27.8          & 28.8           \\
        incorrect\_mapping           & 37.4         & 38.9         & 37.6          & 39.0            & 38.6        & 37.3          & 38.3           \\ \midrule
        \textit{k = 8} \\
        0\_correct                   & 36.7         & 25.3         & 25.6          & 26.2            & 24.3        & 25.8          & 25.3           \\
        25\_correct                  & 40.8         & 29.8         & 30.3          & 30.8            & 27.3        & 29.2          & 28.6           \\
        50\_correct                  & 43.7         & 32.0         & 32.3          & 34.4            & 29.3        & 32.1          & 30.6           \\
        75\_correct                  & 44.0         & 33.6         & 34.8          & 36.4            & 34.2        & 33.5          & 33.4           \\
        gold                         & 44.0         & 35.4         & 34.8          & 36.6            & 34.9        & 35.3          & 35.6           \\
        random                       & 40.5         & 32.2         & 30.9          & 33.8            & 30.8        & 30.8          & 29.8           \\
        incorrect\_mapping           & 37.5         & 41.4         & 40.4          & 41.7            & 38.3        & 39.7          & 39.3           \\ \midrule
        \textit{k = 12} \\
        0\_correct                   & 34.2         & 24.7         & 24.9          & 26.1            & 21.9        & 24.6          & 24.4           \\
        25\_correct                  & 39.5         & 30.0         & 28.4          & 28.4            & 27.0        & 27.3          & 27.7           \\
        50\_correct                  & 43.7         & 35.3         & 31.8          & 34.5            & 30.0        & 31.4          & 32.1           \\
        75\_correct                  & 45.1         & 35.7         & 34.4          & 36.8            & 32.5        & 33.4          & 32.7           \\
        gold                         & 46.8         & 35.8         & 36.7          & 39.6            & 34.6        & 34.8          & 35.3           \\
        random                       & 38.9         & 31.1         & 29.0          & 32.1            & 30.7        & 29.5          & 28.9           \\
        incorrect\_mapping           & 39.5         & 43.1         & 41.5          & 41.7            & 39.4        & 42.6          & 39.6           \\ \midrule
        \textit{k = 16} \\
        0\_correct                   & 34.2         & 24.9         & 24.2          & 25.2            & 22.9        & 26.7          & 24.1           \\
        25\_correct                  & 37.5         & 28.7         & 26.4          & 28.9            & 26.6        & 27.3          & 27.1           \\
        50\_correct                  & 41.6         & 33.1         & 32.0          & 33.4            & 30.6        & 31.2          & 31.3           \\
        75\_correct                  & 45.8         & 37.0         & 36.3          & 39.4            & 32.7        & 34.6          & 33.8           \\
        gold                         & 48.6         & 38.1         & 37.9          & 39.6            & 35.0        & 36.4          & 34.5           \\
        random                       & 38.3         & 30.8         & 28.4          & 33.3            & 30.1        & 30.0          & 29.0           \\
        incorrect\_mapping           & 40.8         & 44.4         & 43.6          & 42.3            & 40.9        & 42.4          & 40.3           \\ \midrule
        \textit{k = 32} \\
        0\_correct                   & 29.0         & 23.9         & 21.8          & 23.5            & 22.9        & 23.8          & 23.8           \\
        25\_correct                  & 37.0         & 27.4         & 24.8          & 29.4            & 27.5        & 27.8          & 29.4           \\
        50\_correct                  & 45.1         & 34.3         & 33.2          & 34.9            & 32.1        & 33.5          & 31.6           \\
        75\_correct                  & 47.8         & 37.2         & 35.9          & 40.2            & 32.9        & 35.0          & 32.9           \\
        gold                         & 49.3         & 38.8         & 37.4          & 39.8            & 36.2        & 36.6          & 34.9           \\
        random                       & 40.8         & 33.7         & 31.3          & 35.8            & 31.7        & 32.1          & 31.4           \\
        incorrect\_mapping           & 44.8         & 46.5         & 44.8          & 42.9            & 41.6        & 42.6          & 41.3           \\ \bottomrule
        \end{tabular}
    }
    \label{table:classification}
    \caption{experimental results of contextual knowledge understanding datasets, reported in Macro-F1.}
\end{table*}
\begin{table*}
    \centering
    \resizebox{\textwidth}{!}{
        \begin{tabular}{lrrrrrrr}
        \toprule
        \textbf{setting}                      & \textbf{Qwen2.5-1.5B} & \textbf{Llama-3.2-1B} & \textbf{gemma-3-1b-pt} & \textbf{\textsc{Hymba-1.5B-Base}} & \textbf{\textsc{Zamba2-1.2B}} & \textbf{mamba-1.4b-hf} & \textbf{mamba2-1.3b-hf} \\ \midrule
        no\_demo                     & 24.8         & 36.1         & 24.2          & 23.1            & 25.0        & 13.9          & 20.5           \\
        \midrule
        \textit{k = 4} \\
        0\_correct                   & 12.3         & 19.5         & 27.0          & 13.2            & 19.4        & 22.1          & 22.0           \\
        25\_correct                  & 45.5         & 44.1         & 51.5          & 32.9            & 39.8        & 44.8          & 42.8           \\
        50\_correct                  & 60.8         & 59.4         & 64.8          & 48.6            & 53.0        & 54.5          & 53.4           \\
        75\_correct                  & 63.4         & 61.9         & 69.1          & 54.1            & 57.0        & 57.2          & 56.2           \\
        gold                         & 64.1         & 66.9         & 70.2          & 58.3            & 59.2        & 58.8          & 58.7           \\
        random                       & 16.9         & 20.1         & 27.1          & 13.7            & 22.2        & 25.7          & 24.2           \\
        \midrule
        \textit{k = 8} \\
        0\_correct                   & 7.7          & 17.2         & 22.2          & 11.3            & 15.9        & 21.5          & 21.2           \\
        25\_correct                  & 33.5         & 37.7         & 47.6          & 29.9            & 36.3        & 40.7          & 39.9           \\
        50\_correct                  & 54.7         & 57.1         & 64.6          & 48.3            & 51.4        & 54.8          & 53.2           \\
        75\_correct                  & 64.8         & 67.2         & 70.8          & 57.6            & 59.4        & 61.0          & 59.2           \\
        gold                         & 67.0         & 68.7         & 72.1          & 59.9            & 61.3        & 61.9          & 60.5           \\
        random                       & 10.3         & 17.1         & 21.9          & 11.9            & 15.7        & 20.5          & 18.8           \\
        \midrule
        \textit{k = 12} \\
        0\_correct                   & 6.0          & 12.9         & 18.7          & 10.8            & 15.7        & 19.7          & 17.9           \\
        25\_correct                  & 41.7         & 42.7         & 49.2          & 32.9            & 39.3        & 48.5          & 42.4           \\
        50\_correct                  & 61.3         & 63.6         & 65.9          & 50.8            & 54.4        & 58.6          & 55.2           \\
        75\_correct                  & 65.1         & 67.8         & 70.9          & 56.1            & 59.8        & 60.7          & 58.9           \\
        gold                         & 67.6         & 70.0         & 72.8          & 60.1            & 62.0        & 62.4          & 60.9           \\
        random                       & 8.4          & 15.5         & 19.1          & 9.5             & 14.9        & 19.7          & 19.9           \\
        \midrule
        \textit{k = 16} \\
        0\_correct                   & 6.6          & 14.7         & 18.7          & 8.6             & 13.1        & 20.9          & 18.7           \\
        25\_correct                  & 40.9         & 41.4         & 48.5          & 34.3            & 40.0        & 47.4          & 42.1           \\
        50\_correct                  & 60.7         & 64.0         & 65.1          & 51.8            & 55.9        & 58.8          & 56.0           \\
        75\_correct                  & 66.1         & 68.7         & 71.3          & 57.3            & 59.8        & 61.3          & 59.3           \\
        gold                         & 68.0         & 69.6         & 72.7          & 60.4            & 62.6        & 62.4          & 60.7           \\
        random                       & 7.5          & 14.7         & 19.4          & 10.9            & 15.1        & 19.7          & 19.3           \\
        \midrule
        \textit{k = 32} \\
        0\_correct                   & 5.2          & 13.2         & 15.6          & 7.1             & 11.6        & 19.3          & 15.8           \\
        25\_correct                  & 40.6         & 46.0         & 43.2          & 32.2            & 42.2        & 49.4          & 41.5           \\
        50\_correct                  & 61.9         & 64.9         & 64.8          & 52.7            & 55.7        & 59.2          & 54.4           \\
        75\_correct                  & 66.9         & 69.0         & 71.2          & 58.1            & 60.9        & 61.7          & 59.5           \\
        gold                         & 68.4         & 69.8         & 72.9          & 61.1            & 63.0        & 62.8          & 61.5           \\
        random                       & 7.7          & 13.5         & 16.7          & 9.1             & 12.7        & 20.4          & 17.6           \\      
        \bottomrule
        \end{tabular}
    }
    \label{table:fvog}
    \caption{Experimental results for parametric knowledge retrieval datasets, reported in Macro-F1.}
\end{table*}
\FloatBarrier

\section{Additional Mechanistic Interpretability Experiment Results}
\label{appendix:additional-interpretability-results}
\begin{figure*}
    \centering
    \includegraphics[width=\linewidth]{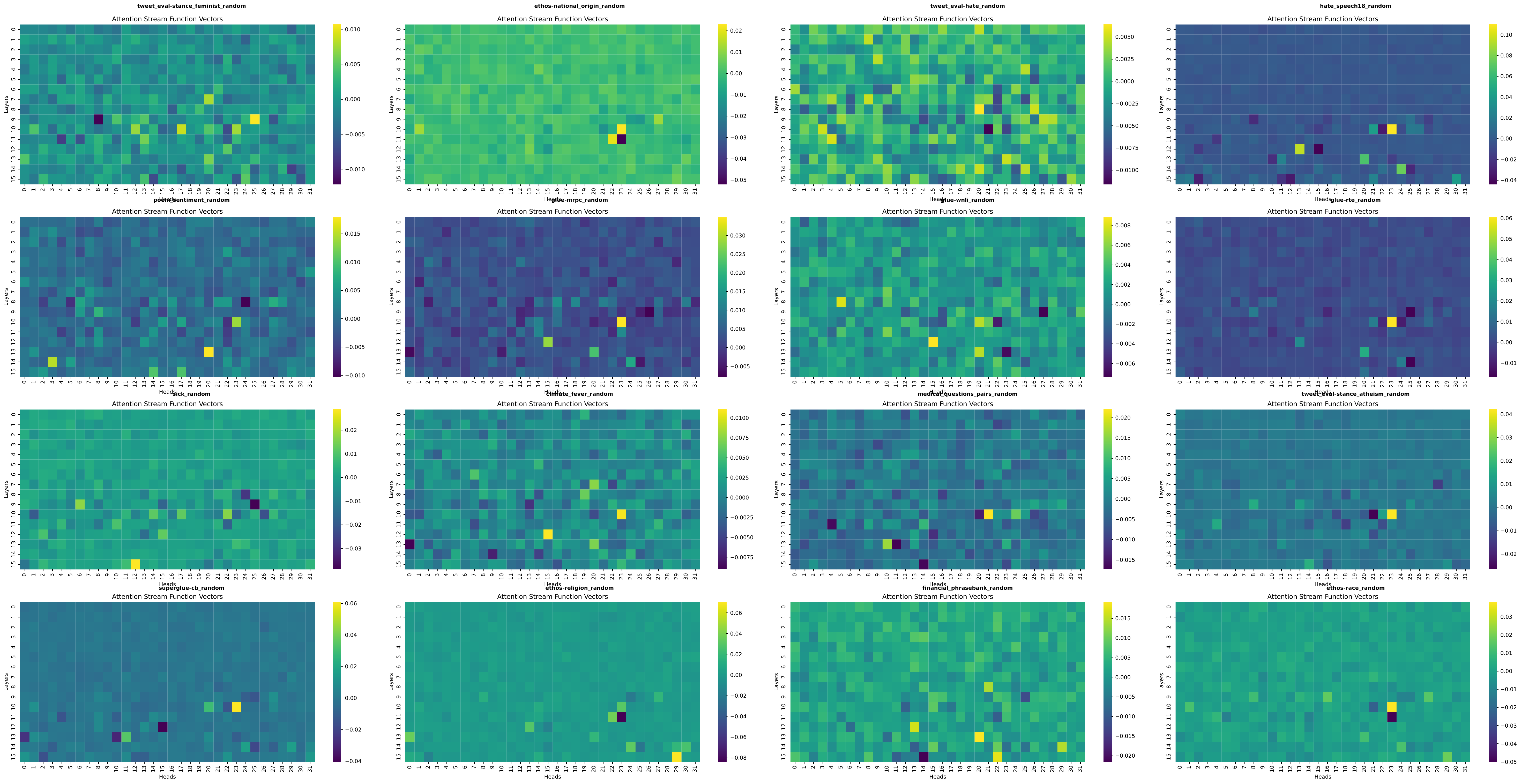}
    \caption{Llama-3.2-1B's AIE heatmaps on contextual knowledge understanding datasets.}
    \label{fig:heatmap-classification-llama}
\end{figure*}
\begin{figure*}
    \centering
    \includegraphics[width=\linewidth]{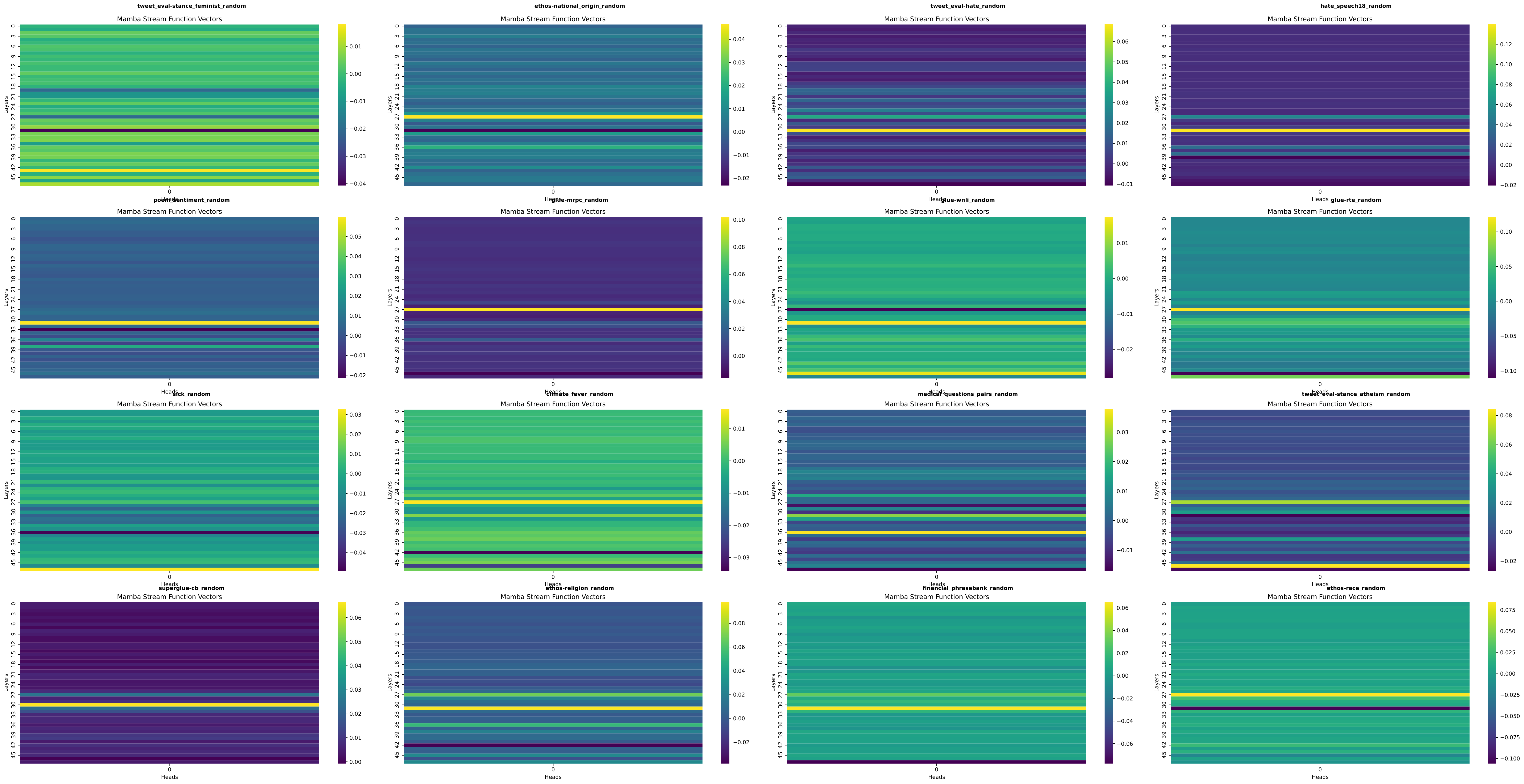}
    \caption{mamba-1.4b-hf's AIE heatmaps on contextual knowledge understanding datasets.}
    \label{fig:heatmap-classification-mamba}
\end{figure*}
\begin{figure*}
    \centering
    \includegraphics[width=\linewidth]{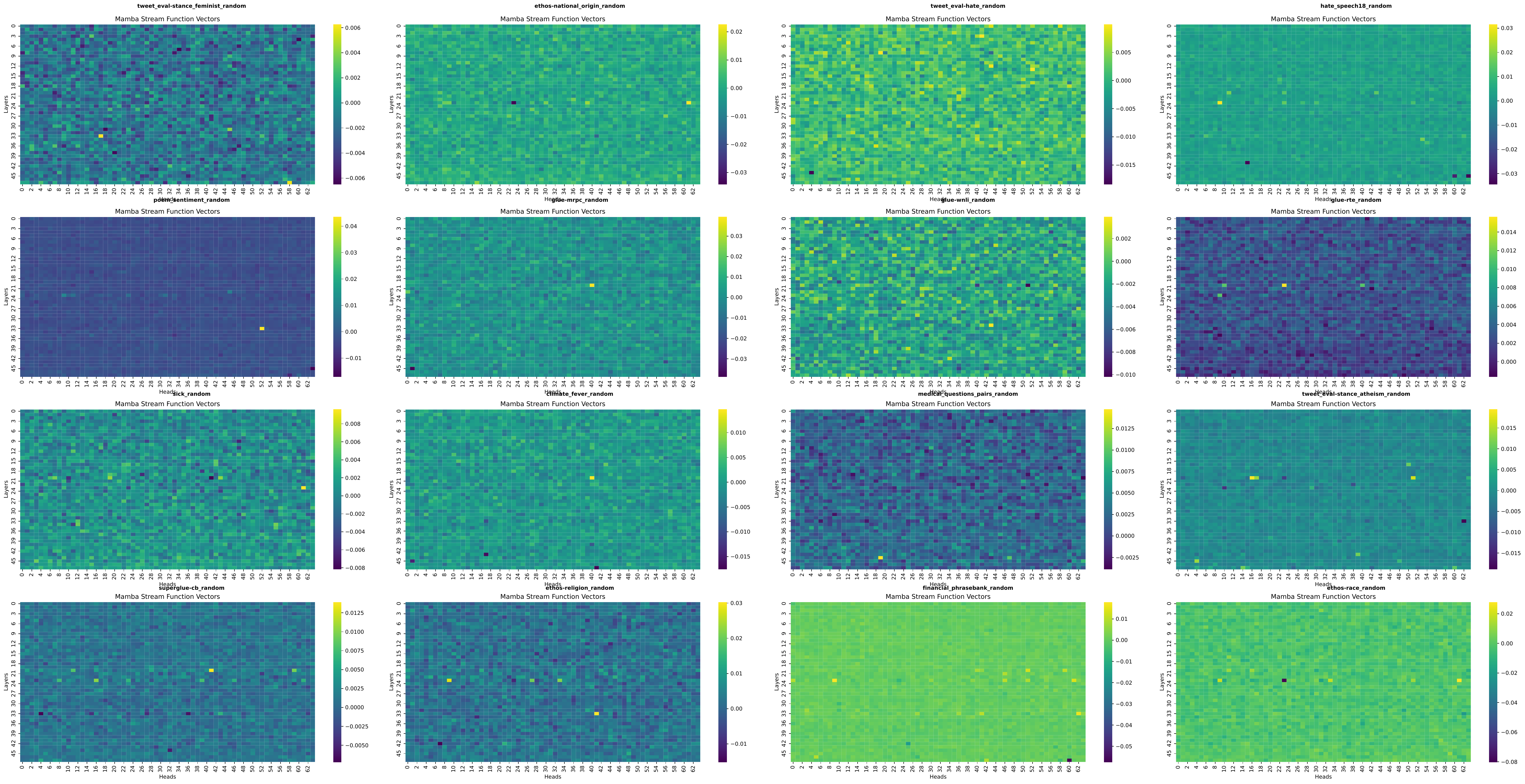}
    \caption{mamba2-1.3b-hf's AIE heatmaps on contextual knowledge understanding datasets.}
    \label{fig:heatmap-classification-mamba2}
\end{figure*}
\begin{figure*}
    \centering
    \includegraphics[width=\linewidth]{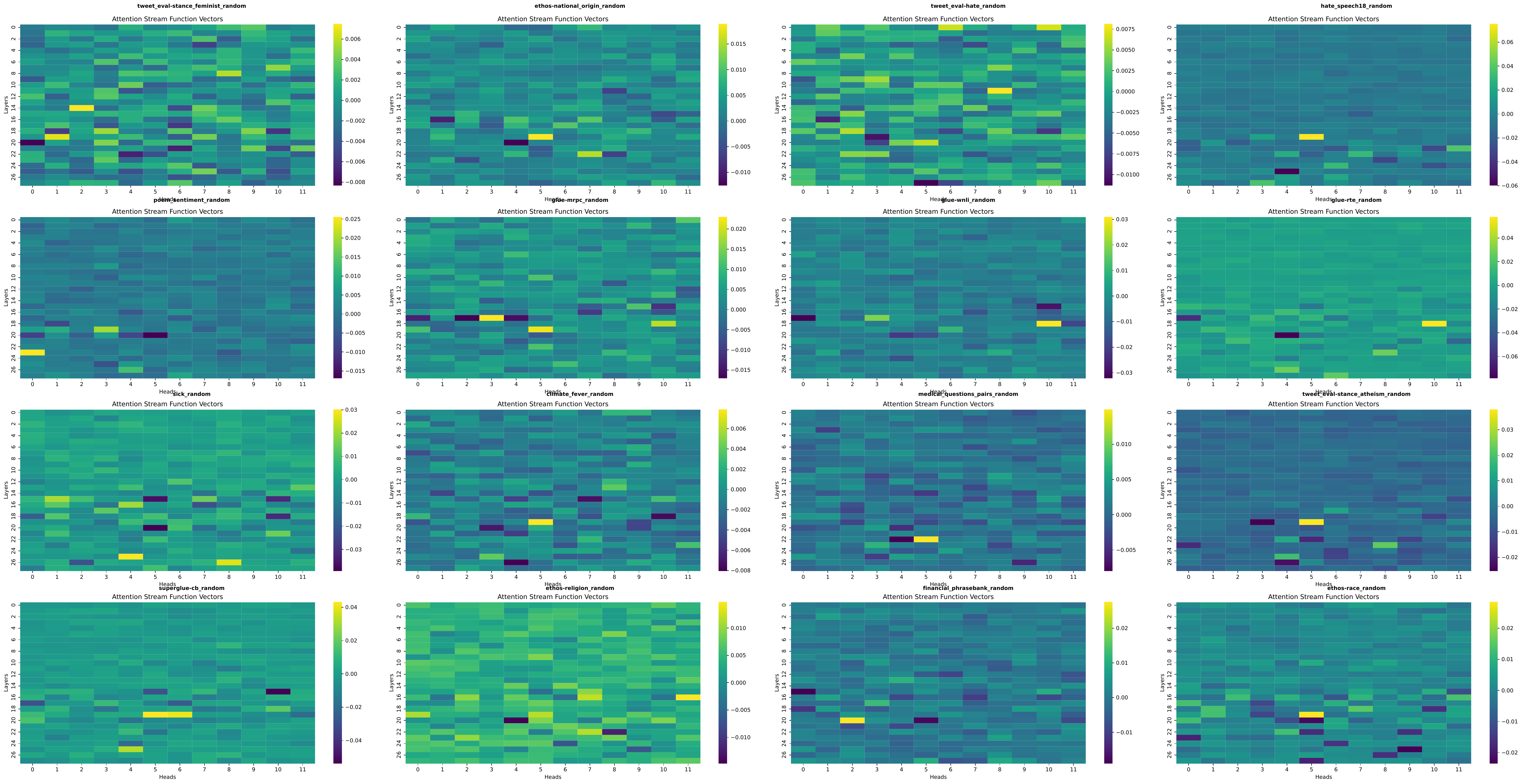}
    \caption{Qwen2.5-1.5B's AIE heatmaps on contextual knowledge understanding datasets.}
    \label{fig:heatmap-classification-qwen}
\end{figure*}
\begin{figure*}
    \centering
    \includegraphics[width=\linewidth]{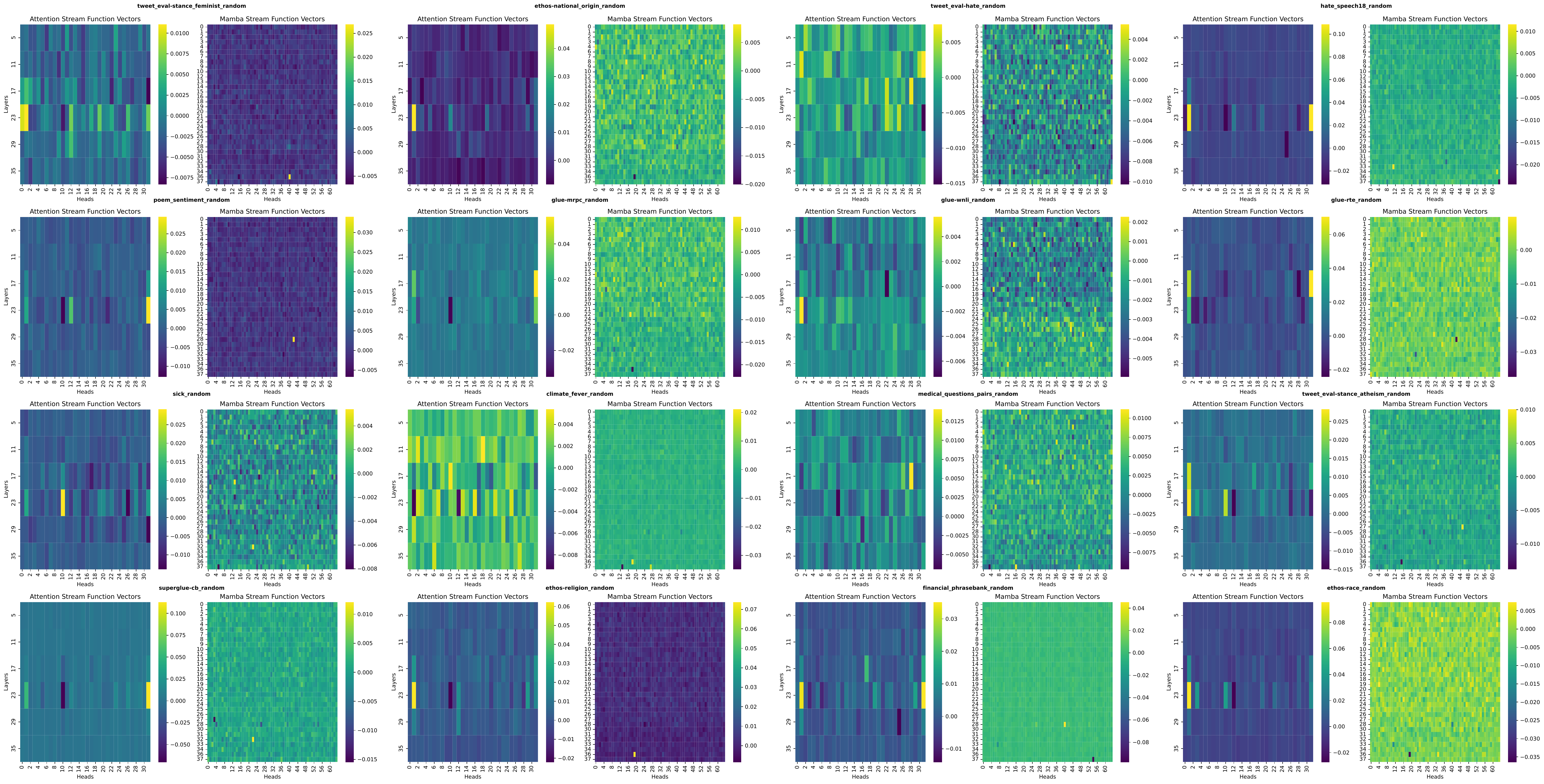}
    \caption{\textsc{Zamba2-1.2B}'s AIE heatmaps on contextual knowledge understanding datasets.}
    \label{fig:heatmap-classification-zamba}
\end{figure*}
\begin{figure*}
    \centering
    \includegraphics[width=\linewidth]{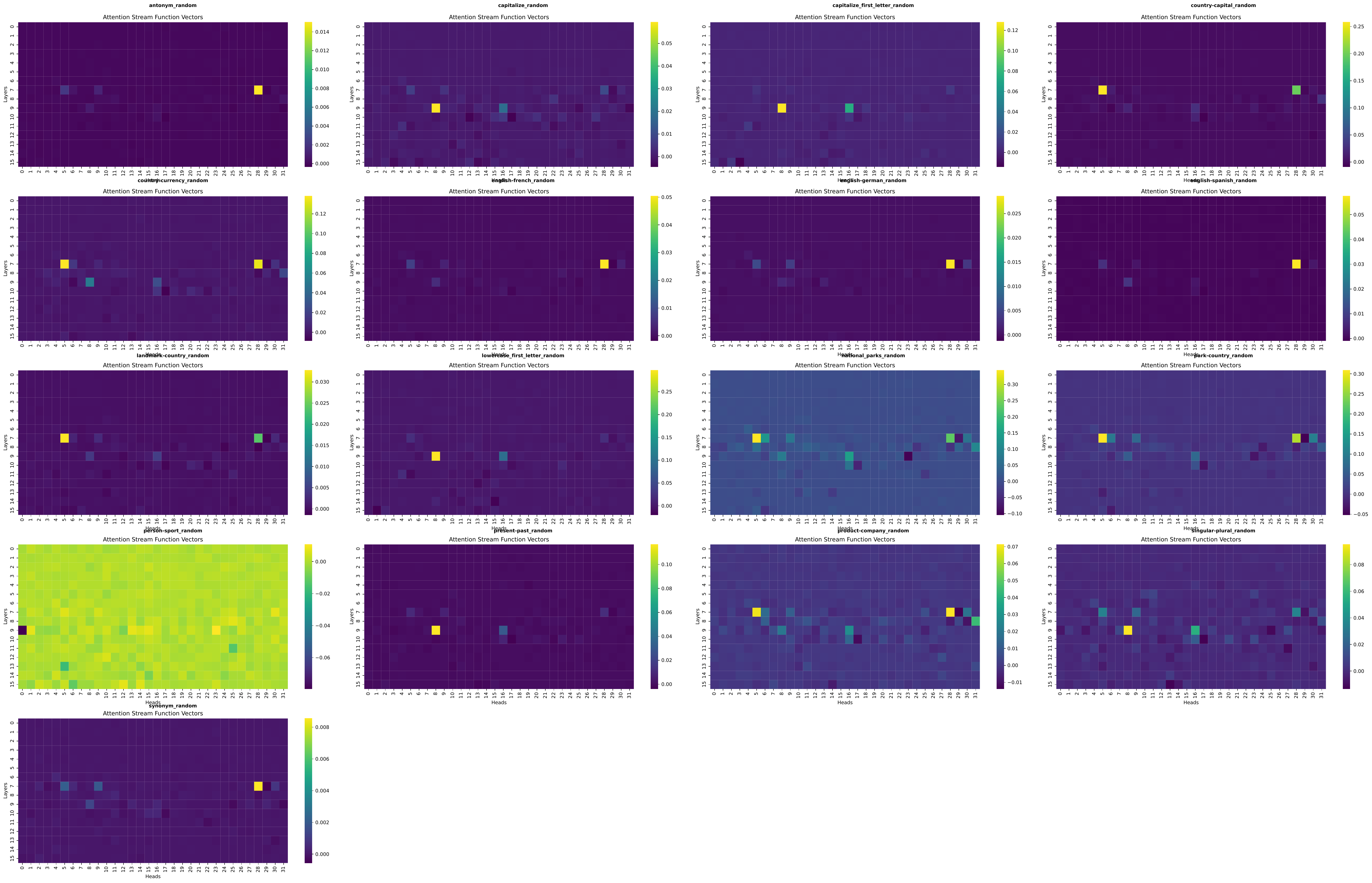}
    \caption{Llama-3.2-1B's AIE heatmaps on parametric knowledge retrieval datasets.}
    \label{fig:heatmap-fv_og-llama}
\end{figure*}
\begin{figure*}
    \centering
    \includegraphics[width=\linewidth]{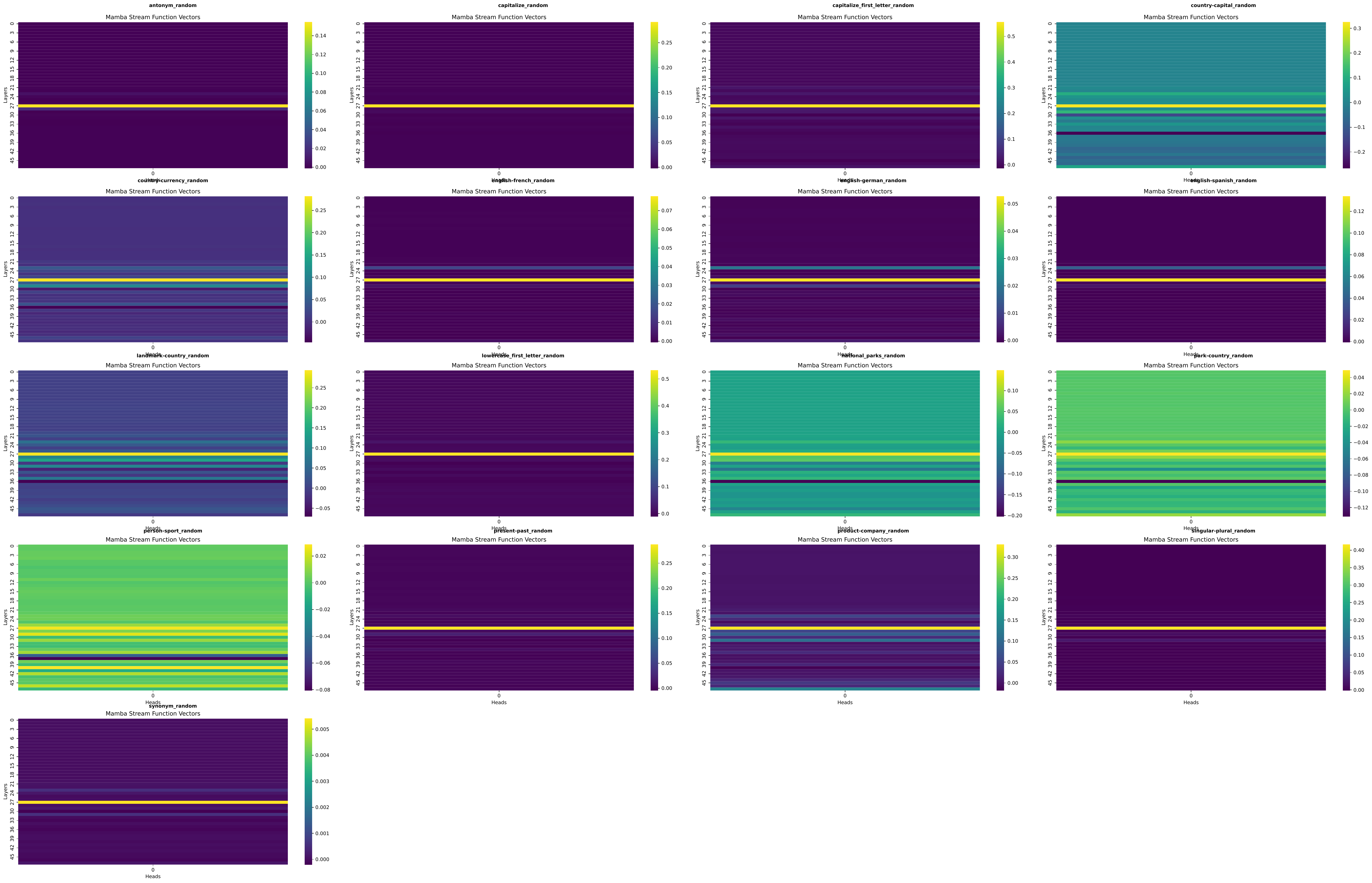}
    \caption{mamba-1.4b-hf's AIE heatmaps on parametric knowledge retrieval datasets.}
    \label{fig:heatmap-fv_og-mamba}
\end{figure*}
\begin{figure*}
    \centering
    \includegraphics[width=\linewidth]{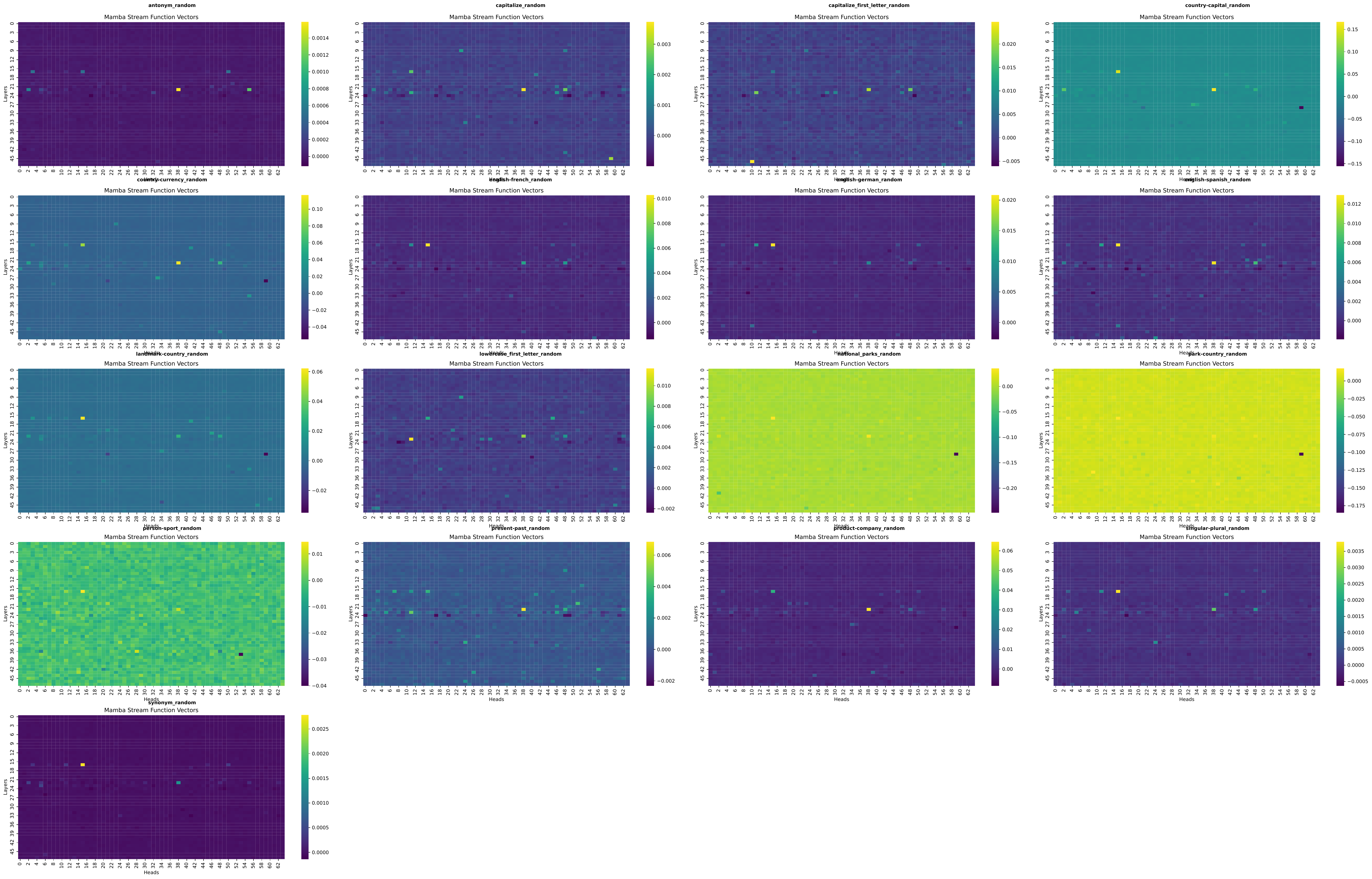}
    \caption{mamba2-1.3b-hf's AIE heatmaps on parametric knowledge retrieval datasets.}
    \label{fig:heatmap-fv_og-mamba2}
\end{figure*}
\begin{figure*}
    \centering
    \includegraphics[width=\linewidth]{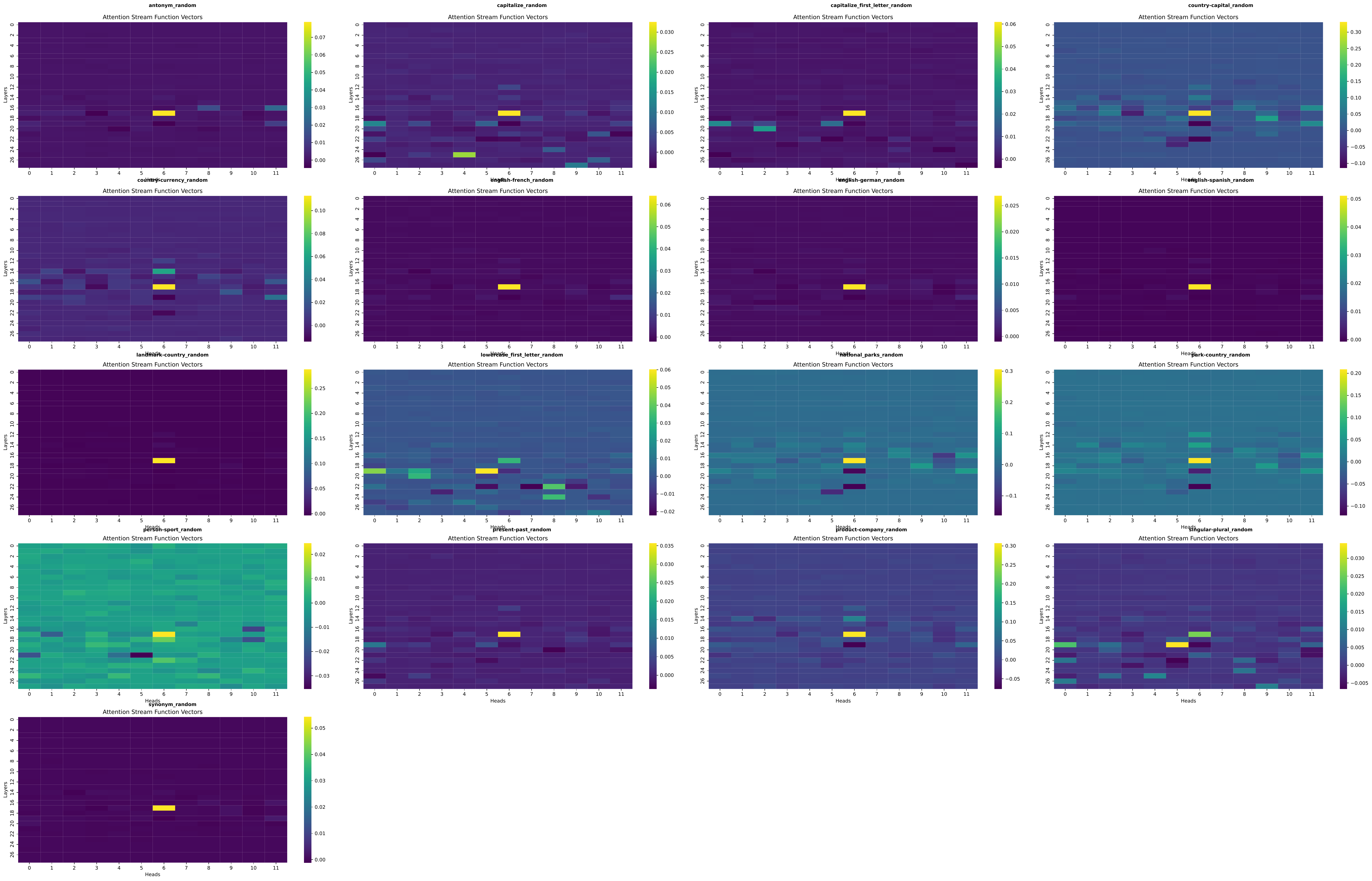}
    \caption{Qwen2.5-1.5B's AIE heatmaps on parametric knowledge retrieval datasets.}
    \label{fig:heatmap-fv_og-qwen}
\end{figure*}
\begin{figure*}
    \centering
    \includegraphics[width=\linewidth]{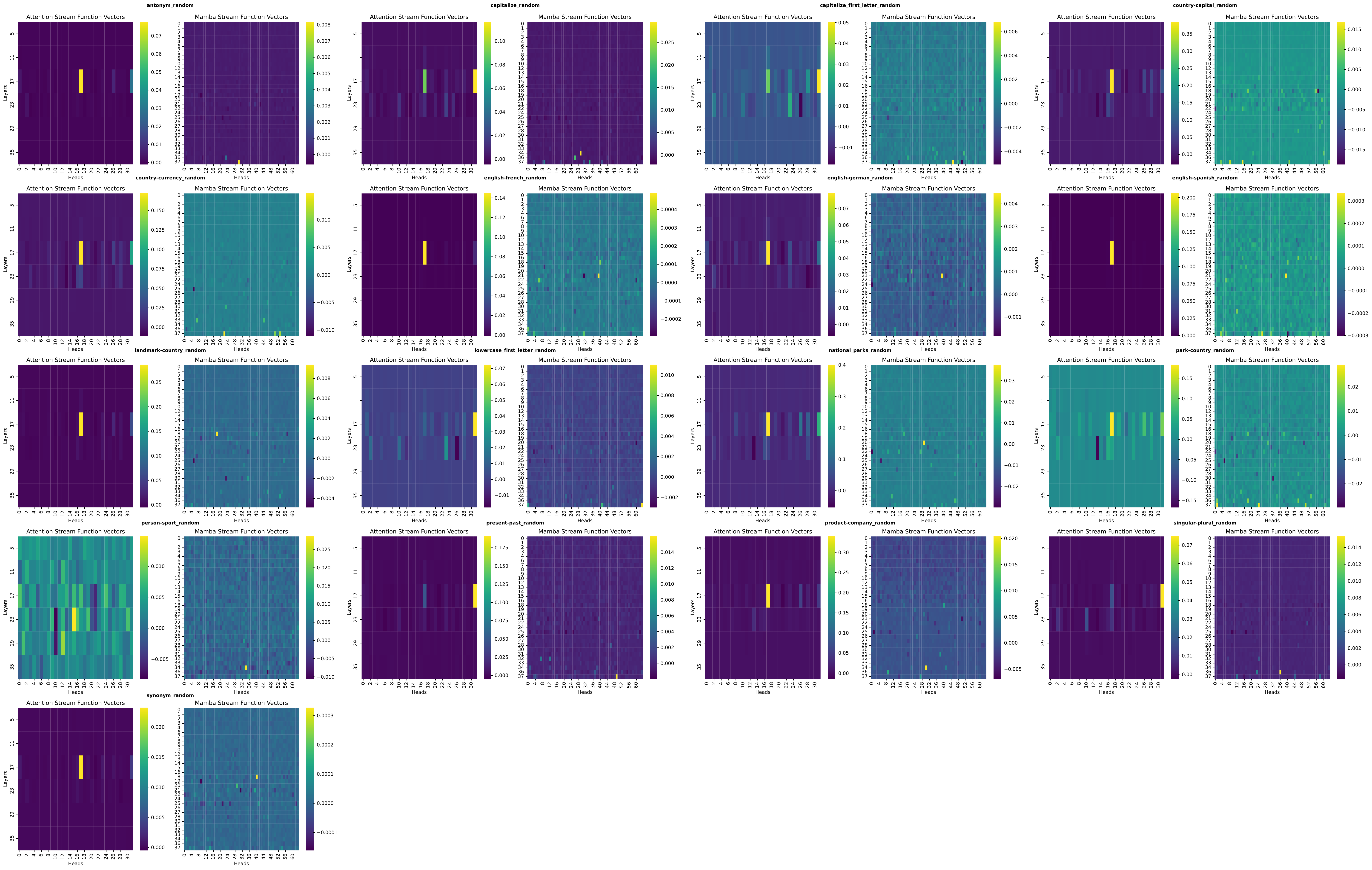}
    \caption{\textsc{Zamba2-1.2B}'s AIE heatmaps on parametric knowledge retrieval datasets.}
    \label{fig:heatmap-fv_og-zamba}
\end{figure*}
\begin{table*}[t]
    \centering
    \small
    \begin{tabular}{ccccccc} \\
        \toprule
        \textbf{top-p} & \textbf{Qwen2.5-1.5B} & \textbf{Llama-3.2-1B} & \textbf{\textsc{Hymba-1.5B-Base}} & \textbf{\textsc{Zamba2-1.2B}} & \textbf{mamba-1.4b-hf} & \textbf{mamba2-1.3b-hf} \\
        \midrule
        0.02 & 0.00\% & 0.00\% & 18.75\% & 18.18\% & -  & 9.84\% \\
        0.04 & 7.69\% & 10.00\% & 12.12\% & 12.36\% & 100.00\% & 10.66\% \\
        0.06 & 20.00\% & 16.67\% & 26.53\% & 14.18\% & 50.00\% & 10.33\% \\
        0.08 & 26.92\% & 15.00\% & 24.24\% & 16.20\% & 33.33\% & 13.06\% \\
        0.10 & 21.21\% & 15.69\% & 31.33\% & 17.41\% & 25.00\% & 13.03\% \\
        0.12 & 20.00\% & 18.03\% & 31.31\% & 16.79\% & 40.00\% & 15.49\% \\
        0.14 & 17.02\% & 19.72\% & 29.31\% & 18.21\% & 33.33\% & 15.81\% \\
        0.16 & 20.75\% & 17.28\% & 28.57\% & 19.27\% & 28.57\% & 17.52\% \\
        0.18 & 21.67\% & 19.57\% & 28.19\% & 20.35\% & 25.00\% & 19.75\% \\
        0.20 & 22.39\% & 18.63\% & 30.12\% & 22.77\% & 33.33\% & 20.85\% \\ 
        \bottomrule
    \end{tabular}
    \caption{Percentage of intersection between top FV heads identified from contextual knowledge understanding datasets and parametric knowledge retrieval datasets. Entries are the percentage of heads in top-p that overlap.}
    \label{table:overlap}
\end{table*}
\begin{figure*}
    \centering
    \includegraphics[width=\linewidth]{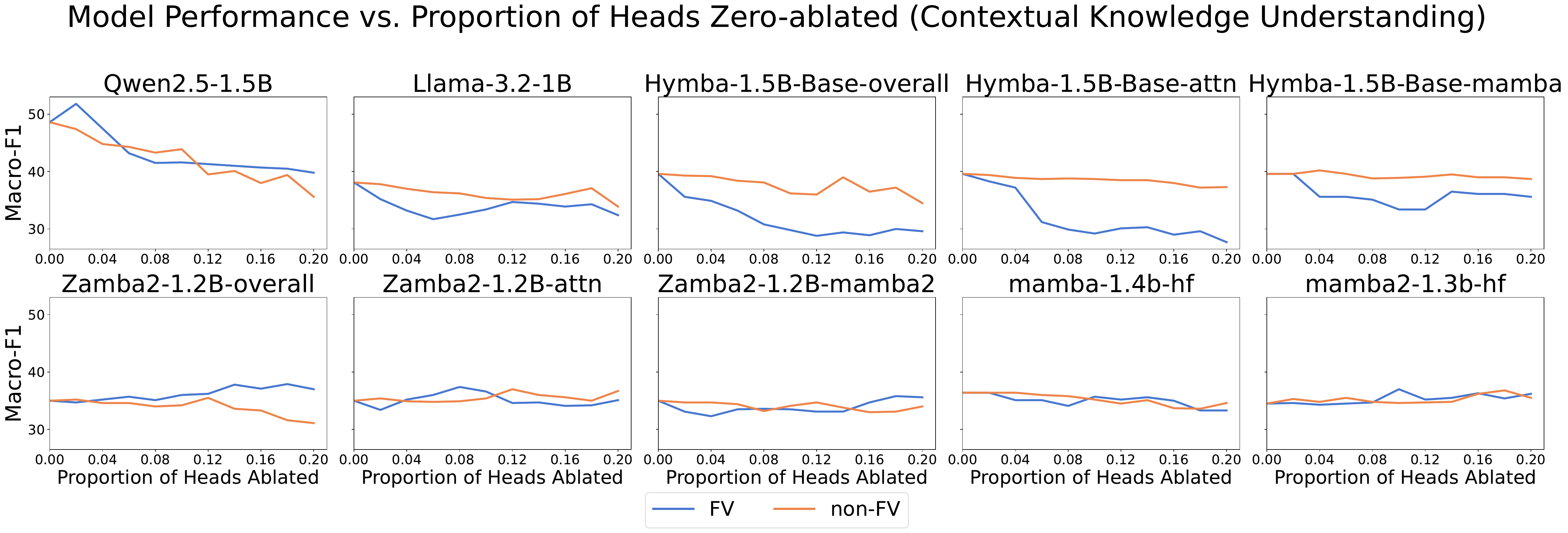}
    \caption{Zero ablation results on contextual knowledge understanding datasets.}
    \label{fig:zero_ablation-classification}
\end{figure*}
\begin{figure*}
    \centering
    \includegraphics[width=\linewidth]{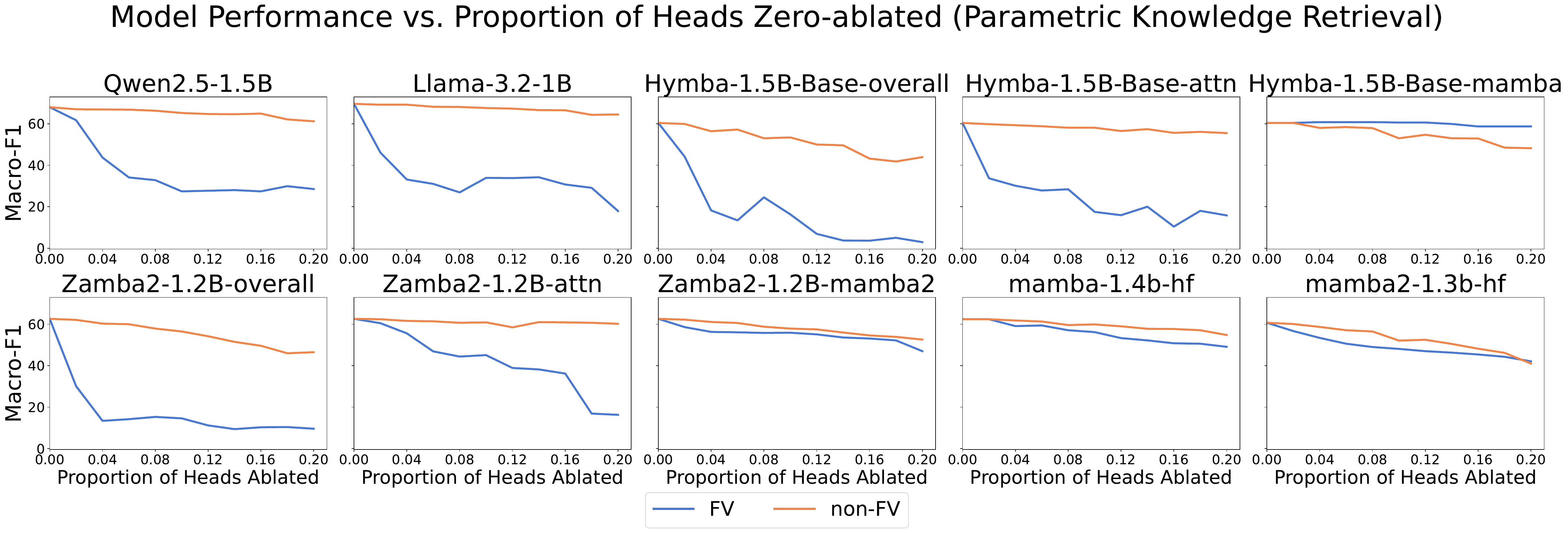}
    \caption{Zero ablation results on parametric knowledge retrieval datasets.}
    \label{fig:zero_ablation-fv_og}
\end{figure*}
\FloatBarrier

\section{Additional Hymba Results}
\label{appendix:additional-hymba-results}
\begin{figure*}[ht]
    \centering
    \includegraphics[width=0.9\linewidth]{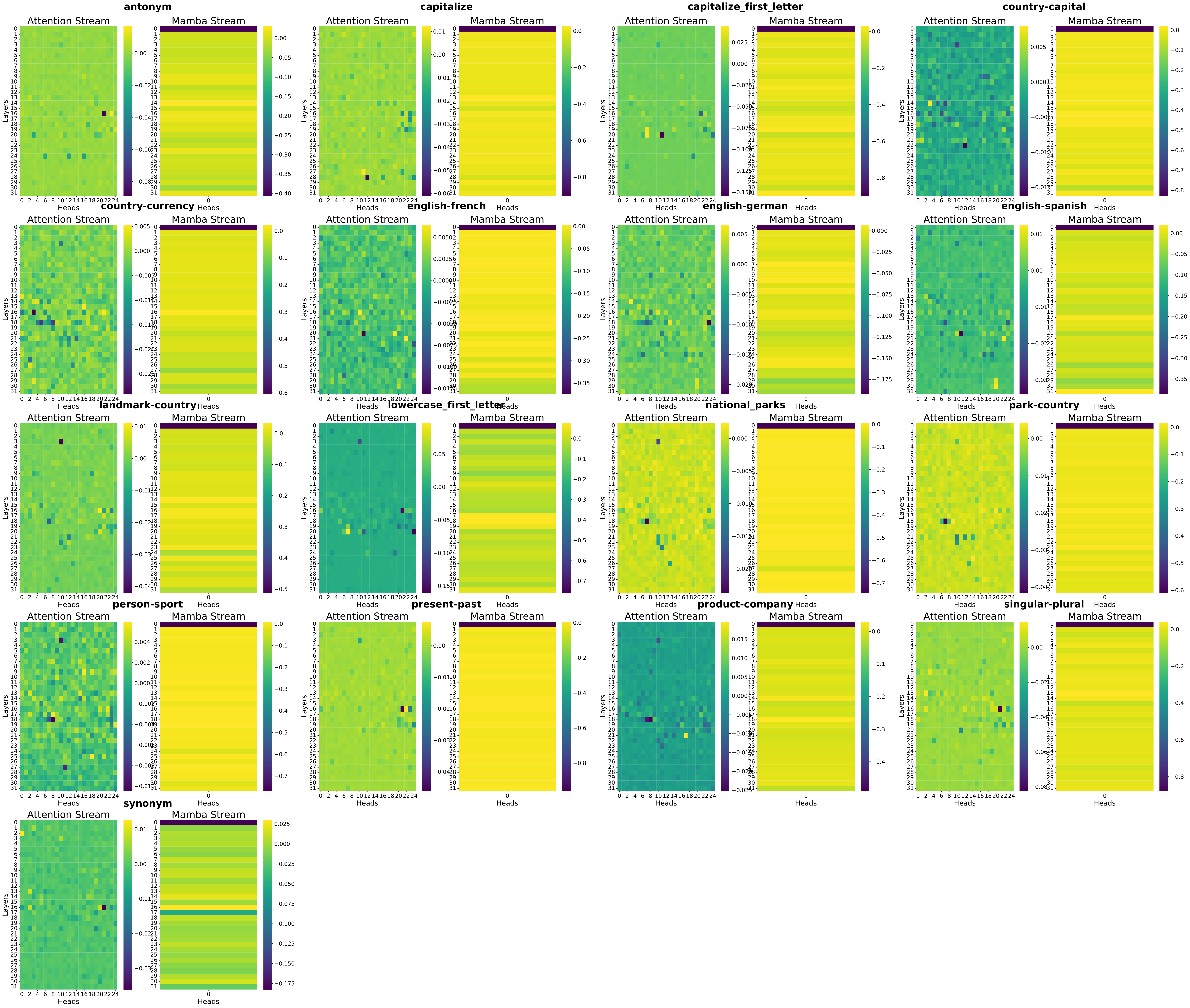}
    \caption{Negative AIE experiment results for \textsc{Hymba-1.5B-Base} on parametric knowledge retrieval datasets. Darker heads are those that caused a greater drop in performance upon removal.}
    \label{fig:hymba-neg-AIE-fvog}
\end{figure*}
\begin{figure*}[ht]
    \centering
    \includegraphics[width=0.9\linewidth]{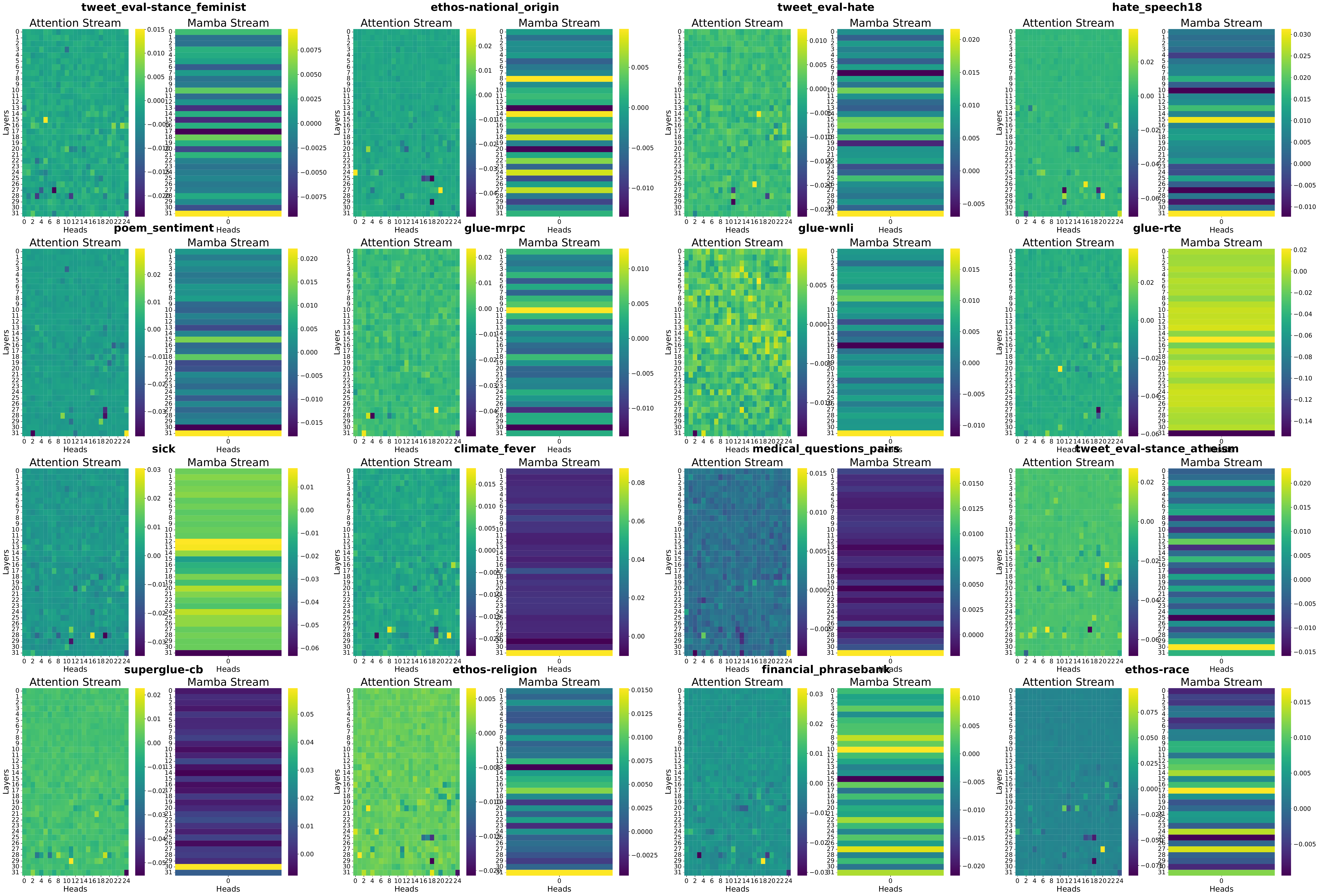}
    \caption{Negative AIE experiment results for \textsc{Hymba-1.5B-Base} on contextual knowledge understanding datasets. Darker heads are those that caused a greater drop in performance upon removal.}
    \label{fig:hymba-neg-AIE-classification}
\end{figure*}
To analyze the less consistent behavior in \textsc{Hymba-1.5B-Base}'s in steering and ablation experiments, we introduce the negative AIE experiment: on the gold-labelled dataset, we mean-ablated each head to observe how much drop in the logit confidence this ablation introduced. We present our results in Figures \ref{fig:hymba-neg-AIE-fvog} and \ref{fig:hymba-neg-AIE-classification}. Surprisingly, we find the results to be extremely consistent on the parametric knowledge retrieval tasks: the Mamba head on the very first layer always cause the most significant drop in performance. Nevertheless, results on the contextual understanding tasks, as opposed to the other, appear to be less consistent. The attention stream seemed to have played a more important role in contextual knowledge understanding, as ablating attention heads caused a more significant drop in AIE magnitude compared to ablating the mamba stream.

\end{document}